\documentclass[runningheads,a4paper]{llncs}

\usepackage{xcolor}
\usepackage{amsmath}
\usepackage{amssymb}
\usepackage{bbm}
\usepackage{hyperref}
\usepackage{graphicx}
\usepackage{booktabs} 
\usepackage{multirow}
\usepackage{mathtools}
\usepackage{xurl}
\usepackage[T1]{fontenc}

\usepackage[backend=biber,style=nature, maxnames=100]{biblatex}
\urldef{\mail}\path|{yuyangw, zijieli, barati}@cmu.edu|    
\newcommand{\keywords}[1]{\par\addvspace\baselineskip
\noindent\keywordname\enspace\ignorespaces#1}

\newcommand{\angstrom}{\mbox{\normalfont\AA}}
\DeclareUnicodeCharacter{2212}{-}
\DeclareUnicodeCharacter{0301}{\'{e}}

\begin{document}
\mainmatter  

\title{Graph Neural Networks for Molecules \\ \normalsize{\textit{A Chapter for Book ``Machine Learning in Molecular Sciences"}}}

\titlerunning{Graph Neural Networks for Molecules}

%
%
\author{Yuyang Wang$^{\text{12}}$ \and Zijie Li$^{\text{1}}$ \and Amir Barati Farimani$^{\text{123}}$\thanks{Corresponding author.}}
\authorrunning{Graph Neural Networks for Molecules}

\institute{
1 Department of Mechanical Engineering, Carnegie Mellon University \\
2 Machine Learning Department, Carnegie Mellon University \\
3 Department of Chemical Engineering, Carnegie Mellon University \\
Pittsburgh, PA 15213, USA \\
\mail}

%
%


\let\oldaddcontentsline\addcontentsline
\def\addcontentsline#1#2#3{}
\maketitle
\def\addcontentsline#1#2#3{\oldaddcontentsline{#1}{#2}{#3}}
\renewcommand{\contentsname}{Contents}


\begin{abstract}
Graph neural networks (GNNs), which are capable of learning representations from graphical data, are naturally suitable for modeling molecular systems. This review introduces GNNs and their various applications for small organic molecules. GNNs rely on message-passing operations, a generic yet powerful framework, to update node features iteratively. Many researches design GNN architectures to effectively learn topological information of 2D molecule graphs as well as geometric information of 3D molecular systems. GNNs have been implemented in a wide variety of molecular applications, including molecular property prediction, molecular scoring and docking, molecular optimization and \textit{de novo} generation, molecular dynamics simulation, etc. Besides, the review also summarizes the recent development of self-supervised learning for molecules with GNNs. 
\end{abstract}

\thispagestyle{empty}
\tableofcontents
\newpage
\pagenumbering{arabic}


\section{Message Passing Graph Neural Networks}
\label{sec:gnn}

\begin{figure}[t]
    \centering
    \includegraphics[width=\linewidth]{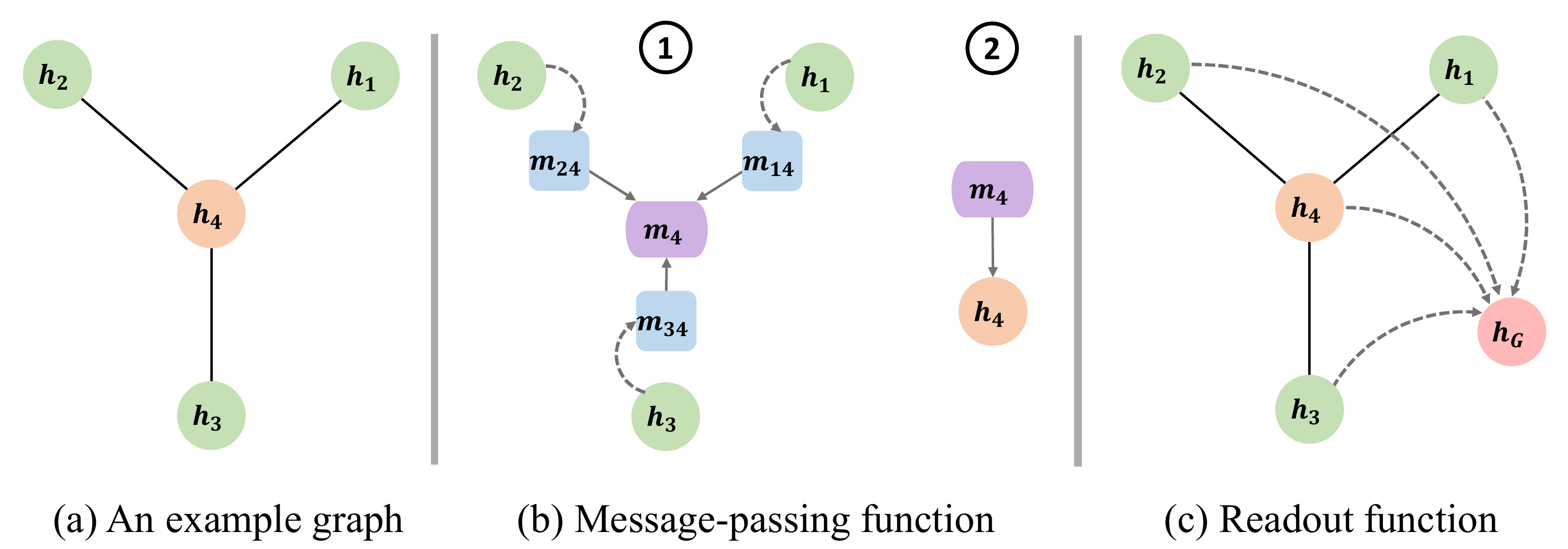}
    \caption{Illustration of (a) an example graph defined by nodes and edges, (b) messaging-passing function containing \textcircled{1} an aggregation operation from $\pmb h_1$, $\pmb h_2$, and $\pmb h_3$ to obtain message $m_4$ and \textcircled{2} an update operation to update the node feature $\pmb h_4$ from $\pmb m_4$, and (c) readout function to obtain the graph feature $\pmb h_G$. }
    \label{fig:gnn}
\end{figure}

Graphs are ubiquitous data structure that expresses a set of objects and the relationships between them. Formally, a graph $\mathcal{G}$ is defined as $\mathcal{G} = (\mathcal{V}, \mathcal{E})$, where $\mathcal{V}$ and $\mathcal{E}$ denotes the set of objects (nodes) and their relationships (edges), respectively \cite{bronstein2017geometric}. Fig.~\ref{fig:gnn}(a) shows an example of a graph containing four nodes and three edges. Graphs are powerful expressions that can model a wide range of systems. For example, the social network can be modeled as a graph where each node represents a person and each edge represents social connections between them, e.g., friendship, spouseship, colleagueship, etc \cite{dai2018learning, fan2019graph, wu2020graph}. Other graphical systems include chemical compounds \cite{duvenaud2015convolutional, kearnes2016molecular}, knowledge graphs \cite{dumontier2014bio2rdf, hamaguchi2017knowledge}, physical systems \cite{sanchez2018graph, thomas2018tensor}, and various other domains \cite{fout2017protein, qi20173d, wang2018nervenet}. Recently, there is growing attention from the machine learning (ML) community to develop ML models, especially deep neural networks (DNNs) \cite{lecun2015deep}, to analyze graphical data \cite{zhou2020graph, wu2020comprehensive}. Previous deep learning models, including convolutional neural networks (CNNs) \cite{krizhevsky2012imagenet, he2016deep} and recurrent neural networks (RNNs) \cite{hochreiter1997long}, fail to directly operate on graphical structures. Graph neural networks (GNNs), a deep learning method, are developed to learn representations from graphs directly \cite{chami2020machine, you2020design}. GNNs have been prevalent in various domains and many different tasks. In what follows, we will introduce the basic concepts and operations of GNNs. 

Modern GNNs are built upon the message-passing layer that aggregates neighboring information to update each node in an iterative manner. The framework is first formalized by Gilmer et al \cite{gilmer2017neural}. Let $\mathcal{G} = (\mathcal{V}, \mathcal{E})$ where the $i$-th node is $u_i \in \mathcal{V}$ and the edge between $u_i$ and $u_j$ is $e_{ij} \in \mathcal{E}$. Each node $u_i$ is initialized as a feature vector $\pmb h_i^{(0)} = \text{Emb}_n(u_i)$ via an node embedding function. Similar to node embedding, each edge $e_{ij}$ is mapped to a feature vector via an edge embedding function $\pmb a_{ij} = \text{Emb}_e(e_{ij})$. The message-passing function is demonstrated in Fig.~\ref{fig:gnn}(b). Each message-passing layer contains two operations: (1) computation and aggregation of the messages from neighboring nodes and edges, and (2) update of the node feature based on the message and old feature. The two operations at $k$-th layers in a GNN is given in Equation~\ref{eq:aggregate} and \ref{eq:combine}, respectively. 
\begin{equation}
    \pmb{m}_{i}^{(k)} = \sum_{u_j \in \mathcal{N}(u_i)} \phi_m^{(k)}(\pmb h_i^{(k-1)}, \pmb h_j^{(k-1)}, \pmb a_{ij}),
    \label{eq:aggregate}
\end{equation} 
\begin{equation}
    \pmb h_i^{(k)} = \phi_f^{(k)} ( \pmb h_i^{(k-1)}, \pmb m_{i}^{(k)}),
    \label{eq:combine}
\end{equation}
where $\pmb h_i^{(k)} \in \mathbb{R}^F$ denotes the feature vector of node $u_i$ at $k$-th layer, $\mathcal{N}(u_i)$ models all the neighbors of node $u_i$. $\phi_m^{(k)}$ and $\phi_u^{(k)}$ are the message and update functions. By conducting the two operations iteratively, node features within the graph are updated and can be utilized for node-level tasks. To conduct graph-level tasks, like molecular property prediction, directly utilizing all the node features might be infeasible as different graphs have different numbers of nodes. To obtain the representation of the whole graph, the readout operation is introduced to down-sample the node features as illustrated in Fig.~\ref{fig:gnn}(c). For a GNN with $K$ layers, after the last message-passing layer, the readout function $R(\cdot)$ is given in Equation~\ref{eq:readout}. 
\begin{equation}
    \pmb h_G = R ( \{\pmb h_i^{(K)} \ | \ u_i \in \mathcal{V}\}). 
    \label{eq:readout}
\end{equation}

The design of the message-passing operation is essential to learning graph representations \cite{battaglia2018relational}. Since proposed, different GNNs have been developed with different aggregation and update functions for expressive graph representation learning \cite{shuman2013emerging, li2015gated, battaglia2016interaction, gao2018large, zhang2018gaan}. Early spectral methods are built upon the spectral representation of graphs. Such methods compute the eigenvectors of the graph Laplacian for aggregating neighboring information and apply nonlinear activation functions on the aggregated feature to update each node \cite{bruna2013spectral, defferrard2016convolutional}. Graph convolutional network (GCN) \cite{kipf2016semi} introduces a simple yet generic GNN framework, where the aggregation is implemented as an element-wise mean pooling over the node and its neighbors, and the update is implemented as a linear transformation $\pmb W$ followed by nonlinear ReLU function \cite{maas2013rectifier} as given in Equation~\ref{eq:gcn}. 
\begin{equation}
    \pmb h_i^{(k)} = \text{ReLU}(\pmb W \cdot \text{MEAN}(\pmb h_j^{(k-1)}) \ | \ {u_j \in \mathcal{N}(u_i) \cup \{u_i\}})).
    \label{eq:gcn}
\end{equation}
GraphSAGE \cite{hamilton2017inductive} formulates aggregation via multiplication of learnable weight matrix and neighboring features followed by ReLU and max-pooling over the aggregated features as given in Equation~\ref{eq:graphsage}.
\begin{equation}
    \pmb h_i^{(k)} = \sigma \left( \pmb W' \cdot \left[ \text{MAX} \left( \text{ReLU} (\pmb W \cdot \pmb h_j^{(k-1)}) \ | \ {u_j \in \mathcal{N}(u_i)} \right) \ \| \ h_i^{(k-1)} \right] \right), 
    \label{eq:graphsage}
\end{equation}
where $[\cdot \| \cdot]$ is the concatenation and $\sigma$ is a nonlinear activation function like sigmoid. The update in GraphSAGE contains a concatenation of the node and aggregated features succeeded by a linear transformation $\pmb W'$. Equation~\ref{eq:gin} shows the message-passing in graph isomorphism network (GIN) \cite{xu2018how}, another widely used GNN architecture. GIN proposes to sum the node features and all the neighboring features and applies a multi-layer perceptron (MLP) to update the node in the message-passing layer. 
\begin{equation}
    \pmb h_i^{(k)} = \text{MLP}( (1+\epsilon) \pmb h_i^{(k-1)} + \sum\nolimits_{u_j \in \mathcal{N}(u_i)} \pmb h_j^{(k-1)}), 
    \label{eq:gin}
\end{equation}
Further, graph attention network (GAT) \cite{velivckovic2017graph} introduces the attention mechanism to message-passing via computing the attention score as the weight coefficient in aggregation. The standard attention follows $\text{Attn}(Q,K,V) = \frac{QK^\top}{\sqrt{d_k}} V$, where $Q$, $K$, and $V$ are quey, key, and value matrices of the embedding of each token, respectively, with the square root of the embedding dimension $d_k$ as the scaling factor. The message-passing in GAT adjusts the attention as given in Equation~\ref{eq:gat} and \ref{eq:gat_attention}. 
\begin{equation}
    \pmb h_j^{(k)} = \sigma(\frac{1}{K} \sum\nolimits_{k=1}^K \sum\nolimits_{u_j \in \mathcal{N}(u_i) \cup \{u_i\} } \alpha_{ik}^k \pmb W^k \pmb h_j^{(k-1)}), 
    \label{eq:gat}
\end{equation}
\begin{equation}
    \alpha_{ij}^k = \frac{\exp(\text{ReLU}(\pmb a^\top [\pmb W^k h_i^{(k-1)} \ \| \ \pmb W^k h_j^{(k-1)}]))}{\sum_{u_j \in \mathcal{N}(u_i)} \exp(\text{ReLU}(\pmb a^\top [\pmb W^k h_i^{(k-1)} \ \| \ \pmb W^k h_j^{(k-1)}]))}, 
    \label{eq:gat_attention}
\end{equation}
where $\alpha_{ij}^k$ is the scalar that measures attention score, $K$ is the number of attention heads, and $\pmb a \in \mathbbm{R}^{2F}$ is the weight vector. Multiple works also leverage RNNs to model the message-passing function. Gated graph neural networks (GCNN) by Li et al. \cite{li2015gated} employs the gated recurrent unit (GRU) \cite{cho2014learning} to aggregate neighboring information and update node features as shown in Equation~\ref{eq:gcnn}.
\begin{equation}
    \pmb h_i^{(k)} = \text{GRU} \left( \pmb h_i^{(k-1)}, \sum\nolimits_{u_j \in \mathcal{N}(u_i)} \pmb W \pmb h_j^{(k-1)} \right).
\label{eq:gcnn}
\end{equation}
Long short-term memory (LSTM) \cite{hochreiter1997long} is implemented similarly as GRU for message-passing functions. Tai et al. \cite{tai2015improved} propose Tree-LSTM that extends LSTM to tree-structured data. Such a model is adapted to general graphs by Peng et al. \cite{peng2017cross} as well as Zayats and Ostendorf \cite{zayats2018conversation}. 

Recently, there are also efforts on adapting the transformer architecture to graphical data \cite{rampavsek2022recipe}. The standard transformer is designed to learn from sequential data like text \cite{vaswani2017attention}. How to encode the structural information of the graph is the major challenge in applying the transformer on graphs. To this end, Graphormer \cite{ying2021transformers} is directly built upon the standard transformer with an effective encoding of the structural information. Specifically, Graphormer includes spatial encoding of the shortest path, edge encoding of edge features on the shortest path, and centrality encoding of the degree centrality of each node. Each encoding is corporated with learnable parameters and added to the attention score. Dwivedi et al. \cite{dwivedi2020generalization} introduce a graph transformer that learns from the local attention of neighboring connections. It also implements the Laplacian eigenvectors of the graph as the positional encoding in place of sinusoidal positional encoding. Recently, tokenized graph transformer (TokenGT) is proposed to demonstrate that pure transformers can be powerful graph representation learners \cite{kim2022pure}. Unlike previous works which only tokenize nodes and integrate edge information in attention updates, TokenGT tokenizes both nodes and edges with orthonormal node identifiers encoding the connectivity of the tokens and trainable type identifiers that encode whether a token is a node or an edge. TokenGT allows adaption of linear attention (e.g., Performer \cite{choromanski2020rethinking}) introduced for pure transformer and reaches $\mathcal{O}(N+M)$ cost, where $N$ and $M$ are the number of nodes and edges, respectively. 

For the readout operation, the primitive choices are pooling over all the node features, including mean-pooling, max-pooling, and summation-pooling. Xu et al. \cite{xu2018how} point out that summation-pooling is more expressive than mean- and max-pooling, which can capture the full multiset while the other two poolings fail. Works have also investigated other readout functions to improve representation learning and computational efficiency. Attention mechanisms have been implemented in place of summation- or mean-pooling as the readout operation \cite{li2015gated, gilmer2017neural}. Vinyals et al. \cite{vinyals2015order} propose set2set function which implements an LSTM for unordered and size-variant input sets. Other methods have probed the rearrangement of the nodes to down-sample the features. Defferrard et al. \cite{defferrard2016convolutional} coarsen the graph into multi-levels via the Graclus algorithm and then rearrange the nodes into a balanced binary tree. The readout is conducted by aggregating the node features in a bottom-up manner. Following the insight, Zhang et al. \cite{zhang2018end} propose SortPool which ranks nodes based on their structural roles within the input graph and truncates the size of the graph after ranking. DiffPool \cite{ying2018diffpool}, on the other hand, develops a differentiable pooling module that generates hierarchical graph representations. It learns a soft clustering assignment at each layer to aggregate node features. SAGPool \cite{lee2019sagpool} combines self-attention with end-to-end hierarchical representation learning which includes both node feature and graph topology. In general, the readout operation can be considered as a special message-passing layer that aggregates all the updated node features within the graph. It also plays an important role in learning expressive graph representations. 


\section{Molecular Graph Neural Networks}
\label{sec:molecular_gnn}

\begin{figure}[t]
    \centering
    \includegraphics[width=0.8\linewidth]{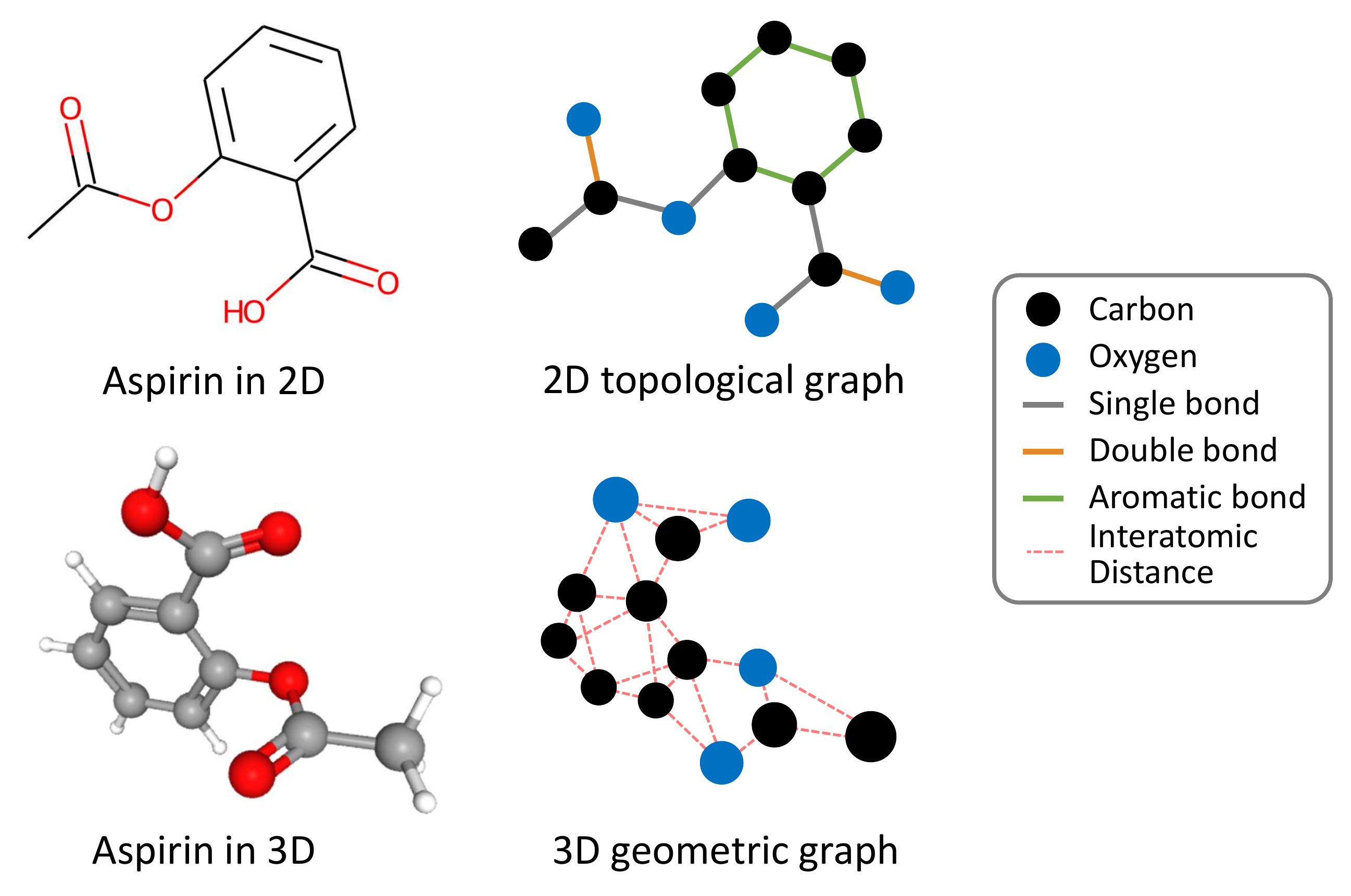}
    \caption{Illustration of building graphs from molecules using 2D topological and 3D geometric information. The aspirin molecule is shown as an example. }
    \label{fig:mol_graph}
\end{figure}

The previous section has introduced the basic concepts of the message-passing GNN and its prevalent variants. This section focuses on the GNN architectures that are adapted and designed for molecular representation learning. As the focus of this chapter, molecules, can be naturally viewed as graphs \cite{atz2021geometric}. Fig.~\ref{fig:mol_graph} illustrates examples of how molecule graphs are built. Within the graph, each node models an atom, and each edge models the interatomic interactions. Interactions include two-dimensional (2D) topological information like covalent bonds and 3D geometric information like distances and angles. Graph representations have demonstrated advantages over other molecular featurization techniques. Some methods develop a language that converts a molecule into a one-dimensional (1D) string, e.g., simplified molecular-input line-entry system (SMILES) \cite{weininger1988smiles}, SMILES arbitrary target specification (SMARTS) \cite{smarts}, and recently developed self-referencing embedded strings (SELFIES) \cite{krenn2020selfies}. Other works propose rules that can embed each molecule into a feature vector, namely a molecular fingerprint or a descriptor, that usually encodes the presence or absence of certain substructures in the molecule. Examples include extended-connectivity fingerprint (ECFP) \cite{rogers2010extended} and molecular access system (MACCS) keys \cite{durant2002reoptimization}. Both string-based and fingerprint-based methods have been widely and successfully implemented in many applications including molecular similarity search, clustering, and virtual screening. With the rise of deep learning in recent years, deep learning methods have been applied to learn molecular representations from molecular languages and fingerprints. Though these featurization techniques follow certain rules to encode substructure information, they, however, still struggle to directly model the topological and geometric information of molecules. GNNs, which learn representation from molecule graphs, are capable of encoding the graphical structures. Many works have investigated the GNN architectures for learning from molecule graphs. Also, atomic and bond descriptors, like atomic number, charge, chirality, hybridization, bond type, etc., can be added to the node and edge features. This chapter focuses on the message-passing architecture design for molecule graphs. 

Duvenaud et al. \cite{duvenaud2015convolutional} introduce one of the pioneering works that learning molecular representations via a GNN. In this work, node features are aggregated through concatenation and are updated through a learnable matrix conditioned on the node degree followed by a sigmoid function. The readout operation computes a weighted summation over node features from all the layers. Another trailblazing model, named Weave, from Kearnes et al. \cite{kearnes2016molecular} proposes molecular graph convolutions that update edge features in each message-passing layer. Specifically, at each layer, one first updates the edge features from connected nodes and then updates the node features by aggregating the edges. Hu et al. \cite{Hu2020Strategies} extend the standard GIN by including the edge features in message-passing as given in Equation~\ref{eq:gine}. 
\begin{equation}
    h_v^{(k)} = \text{MLP} \left( h_v^{l-1} + \sum\nolimits_{u_j \in \mathcal{N}(u_i)} (h_j^{(k-1)} + a_{ij}) \right), 
    \label{eq:gine}
\end{equation}
where $a_{ij}$ is the embedding of the edge between nodes $u_i$ and $u_j$, containing the information of bond type and direction. 
Glimer et al. \cite{gilmer2017neural} elucidate a simple yet unified GNN framework named message passing neural network (MPNN) which most of the previous models fall into. They then propose enn-s2s, a variant of MPNN that aggregates neighboring information as given in Equation~\ref{eq:mpnn}. 
\begin{equation}
    a_i^{(k)} = \sum_{u_j \in \mathcal{N}(u_i)} \text{MLP}(a_{ij}) h_j.
\label{eq:mpnn}
\end{equation}
Besides, it follows set2set \cite{vinyals2015order} as the readout function to obtain molecule representations. Yang et al. \cite{yang2019analyzing} further proposed directed MPNN (D-MPNN) that uses information associated with directed edges instead of information associated with vertices in standard MPNN. In addition, D-MPNN implements rich atom and bond features to improve the expressiveness of molecule graphs. Attention mechanisms have been developed for molecular GNNs. Xiong et al. \cite{xiong2019pushing} introduce AttentiveFP containing atom embedding layers and molecule embedding layers. The atom embedding layers borrow the attention mechanism introduced in GAT to aggregate local messages and the output attention context is fed into a gated recurrent unit (GRU) together with atom features from the previous layer to obtain updated context. The molecule embedding layers assume a virtual node that connects to all atoms and follows the same pattern as atom embedding layers. The final output of the virtual node is used as the representation of the whole molecule graph. Not only the attention mechanisms but also the transformer architectures are investigated for molecule graphs. Rong et al. \cite{rong2020self} introduce GTransformer which combines the message-passing framework with transformer. GTransformer applies a bi-level message-passing strategy that aggregates and updates the information on both nodes and edges. Besides, it employs the residue connection and dynamic message-passing by randomly choosing the number of aggregation hops. 

So far, we have discussed GNNs that model molecules as 2D graphs. Namely, these models only consider topological information while ignoring geometric information like distances and angles in the 3D Euclidean space. Such 2D GNNs have demonstrated promising performance in many applications. However, 3D information is crucial as it is closely related to the energy landscape and molecules rely on 3D conformation to function in practice. Nevertheless, simply adding the positional coordinates into GNNs can be problematic as translation or rotation of the molecule will change the output of the models. One would expect to design GNN architectures whose output is immutable to the rotations and translations of 3D molecular structures \cite{han2022geometrically}. To formalize, we borrow the concept from group theory and denote the set of proper rigid transformations (i.e., translations and rotations) in n-dimensional Euclidean space as SE(n) \cite{blanco2021tutorial}. Many works thus explore SE(3)-invariant GNNs for molecules existing in 3D Euclidean space. Sch{\"u}tt et al. \cite{schutt2017quantum} introduce the deep tensor neural network (DTNN) that model the distances between atoms in message-passing. DTNN extract the graph-level representation by feeding each node feature into an MLP and summing them up. Following DTNN, Sch{\"u}tt et al. further propose SchNet \cite{schutt2017schnet} that is composed of well-designed layers to model local correlations between atoms. Equation~\ref{eq:schnet_atom} elaborate the atom-wise update while Equation~\ref{eq:schnet_interact} and \ref{eq:schnet_cfconv} elaborate continuous-filter convolution at the $(k+1)$-th layer in SchNet. 
\begin{equation}
    h_i^{(k)} = W h_i^{(k)} + b
\label{eq:schnet_atom}
\end{equation}
\begin{equation}
    h_i^{(k)} = \sum\nolimits_{u_j \in \mathcal{N}(u_i)} h_j^{(k)} \circ W^{(k-1)}_{\text{cf}}, 
\label{eq:schnet_interact}
\end{equation}
\begin{equation}
    W^{(k)}_{\text{cf}} = \text{SoftPlus}(W_2^{(k)} \cdot \text{SoftPlus} (W_1^{(k)} \cdot \big|\big|_{u_j \in \mathcal{N}(u_i)} \text{RBF}(d_{ij}) ) ),
\label{eq:schnet_cfconv}
\end{equation}
where $\circ$ is an element-wise multiplication, $d_{ij}$ is the distance in Euclidean space between nodes $u_i$ and $u_j$, and $\text{RBF}(d_{ij}) = \big|\big|_{k=1}^K \exp(-\gamma \| d_{ij} - \mu_k \|^2)$ concatenates radial basis functions with $\mu_k = 0.1k \angstrom$ and $\gamma = 10\angstrom$ that expands the interatomic distance from a scalar to a vector of $\mathbb{R}^K$. Besides, it implements SoftPlus, a smooth approximation to ReLU, as nonlinear activation functions. Each message-passing layer starts with an atomise-wise update followed by a continuous-filter convolution operation. It then conducted two atom-wise updates with a SoftPlus activation in between to obtain the combination term $v_i^{(k)}$. The final output node feature at the $(k+1)$-th layer is updated by $h_i^{(k+1)} = h_i^{(k)} + v_i^{(k)}$. SchNet effectively encodes 3D distance information to molecular GNN and inspires many follow-up works in this domain. PhysNet by Unke et al. \cite{unke2019physnet} also leverages the interatomic distances to build an SE(3)-invariant GNN, which adapts the interaction blocks to update node features from distances and the residual blocks to learn representations with deeper neural networks. DimeNet by Gasteiger et al. \cite{gasteiger2019directional} follows the architecture of PhysNet while integrating additional angular information expanded with Fourier-Bessel representations. The message-passing at the $k$-th layer from node $u_j$ to $u_i$ takes in not only the distance $d_{ij}$ but also angles $\angle u_k u_j u_i$ as well as $h_k^{(k-1)}$, where $u_k \in \mathcal{N}(u_j) \backslash \{u_i\}$. The same team further proposes an improved version named DimeNet++ \cite{klicpera2020fast} with fast interactions and embedding hierarchy. Fang et al. \cite{fang2022geometry} report GeoGNN containing the atom-bond graph and bond-angle graph to incorporate both interatomic distances and angles. In the atom-bond graph, each node represents an atom and each edge represents a covalent bond, while in the bond-angle graph, each node represents an atom-pair and each edge represents the angle between two pairs. GeoGNN builds the message-passing function based on GIN \cite{xu2018how} and updates the atom-bond and bond-angle graphs iteratively. Adams et al. \cite{adams2021learning} introduce a model that is invariant to rotations of rotatable bonds. 

Though SE(3)-invariant has merits for GNN in various molecular property predictions, it is still limited for expressing graph representations in some aspects. One limitation is that such invariance requires the message-passing to contain only features of distances or angles but is unable to encode directional information. Besides, for applications like force field prediction where the output is expected as a set of 3D vectors, SE(3)-invariant GNNs give the same output for molecules with different rotations in 3D. However, the force field is expected to rotate together with the rotation of the molecular system. To this end, SE(3)-equivariance is introduced which generalizes the concept of invariance. Formally, equivariance is a kind of symmetry for function. For function $\Phi: \mathcal{X} \rightarrow \mathcal{Y}$ (e.g., $\Phi$ can be a deep neural network), it is equivariant with respect to group $G$ such that it commutes with any group actions $g \in G$ on $\mathcal{X}$ and $\mathcal{Y}$ as shown in Equation~\ref{eq:equivariance}. 
\begin{equation}
    \rho_g^{\mathcal{Y}} (\Phi (x)) = \Phi (\rho_g^{\mathcal{X}}(x)), \forall x \in \mathcal{X}, g \in G, 
\label{eq:equivariance}
\end{equation}
where $\rho^{\mathcal{X}}(g)$ and $\rho^{\mathcal{Y}}(g)$ are the group representation of group action $g$ on $\mathcal{X}$ and $\mathcal{Y}$ space, respectively. The group representation of group action $g \in G$ on vector space $\mathcal{V}$ is defined as: 
\begin{equation}
    \rho: G \mapsto GL(V),
\label{eq:group repr}
\end{equation}
such that $\rho(g_1g_2)=\rho(g_1)\rho(g_2), \forall g_1, g_2 \in G$ (function satisfies this property is called group homomorphism). The group representation allows operating abstract mathematical object-group action, on the vector space of particular interest. More specifically, for 3D molecules' modeling, we are interested in studying the equivariance with respect to 3D rotations/translations, which corresponds to the 3D special Euclidean group, SE(3). A useful property of the equivariant functions is that composing them will yield another equivariant function, which means the whole network is equivariant if each layer is equivariant. In addition, translation invariance (invariance is also equivariance) is straightforward to achieve in most neural networks as long as they do not use global coordinates as features. Therefore in most of the models, the 3D rotation group SO(3) is the only group that requires special care.

The first major category of equivariant networks is based on irreducible representation, tensor product, and spherical harmonics.
For 3D rotations, its representations are orthogonal matrices and can always be decomposed into the irreducible representation of the following form:
\begin{equation}
    \rho(g)=Q^T\left[\bigoplus_{l} D^{(l)}(g)\right]Q,
\label{eq:wigner d}
\end{equation}
where $Q$ is an $N\times N$ orthogonal matrix (for change of basis), $\bigoplus$ is the direct sum, and $D^{(l)}(g)$ is the \textit{Wigner D-matrices} for group action $g$ \cite{gilmore2008}. Vectors transformed under $D^{(l)}(g)$ are called $l$-th order vectors, which have a length of $2l + 1$. Based on these representations, useful learnable equivariant layers can be developed. A common recipe is first building learnable filters with spherical harmonics and then composing them with input features through the tensor product, which ascribes to the fact that spherical harmonics are the equivariant basis for SO(3) and the tensor product is equivariant. Moreover, the tensor product of the irreducible representations of the vector space can be evaluated by looking up a set of pre-calculated coefficients, which are called Clebsch-Gordon coefficients.

Tensor Field Network \cite{thomas2018tensor} proposes a SO(3) equivariant layer under this recipe:
\begin{equation}
h^l_{\text{out}, i} = W^{ll}h^l_{\text{in}, i} + \sum_{k\geq 0}\sum_{j\in \mathcal{N}(i)}^n W^{lk}(\vec{r}_j-\vec{r}_i)h^k_{\text{in}, j},
\label{eq:tfn layer}
\end{equation}
where $h^l_i$ denotes the $l$-th order vector of input/output features, $W^{lk}(\vec{r}_j-\vec{r}_i)$ is a learnable filter conditioned on relative position $\vec{r}_j-\vec{r}_i$. More specifically, the learnable filter is derived by multiplying a learnable scalar with Clebsch Gordon coefficients and spherical harmonic basis function:
\begin{equation}
W^{lk}(\vec{r}_j-\vec{r}_i) = \sum_{J=|k-l|}^{k+l}\psi_J^{lk}\left(||r_{ij}||\right) \sum_{m=-J}^{J} Y^{(J)}_{m} \left(\frac{\vec{r}_j-\vec{r}_i}{||r_{ij}|| }\right) Q_{Jm}^{lk},
\end{equation}
where $\psi_J^{lk}(\cdot)$ is a learnable function conditioned on the interatomic distance: $||r_{ij}||=||\vec{r}_j-\vec{r}_i||_2$, $Y^{(J)}_m$ denotes the $m$-th dimension of $J$-th spherical harmonics, and $Q_{Jm}^{lk}$ denotes the $(J, m)$-th coefficient in Clebsch-Gordon matrice. The composed learnable filter $W^{lk}(\vec{r}_j-\vec{r}_i) $ with a shape $(2l+1)\times(2k+1)$ will map type-$k$ features to type-$l$ features. Tensor Field Network, there is an active line of works \cite{fuchs2020se, brandstetter2022geometric, cormorant} in using operation on irreducible representation to build SE(3) equivariant network. 

Another direction to build an equivariant network is to exploit the properties of vectorial features instead of leveraging tools from group representation theory. The principle is that the operation on the directional features (e.g. velocities) can only be linear. Therefore, when building the message passing layer, the operations on vector features $\vec h$ are restricted to: (a) Linear projection $W \vec h$; (b) Dot product: $<\vec h_1, \vec h_2>$; (c) Tensor product: $\vec h_1 \otimes \vec h_2$. An example of a such recipe is PaiNN \cite{paiNN2021}, where it proposes an equivariant message passing block by extending the continuous and invariant message passing layer proposed in \cite{schutt2017schnet}. Each layer in the PaiNN comprises two sub-blocks, a message passing block, and an update block. Inside the message passing block, the update of scalar features $\Delta s_i$ are calculated as:
\begin{equation}
    \Delta s_i = \sum_{j \in \mathcal{N}(i)} \phi_s(s_j) \odot W_s(||r_{ij}||), 
\end{equation}
where $\odot$ denotes Hadamard product, $\phi_s(\cdot)$ is a learnable atom-wise  function, $W_s(||r_{ij}||)$ is the learnable continuous filter conditioned on interatomic distance, which is a linear combination of the radial basis function (PaiNN adopts the radial basis function proposed in \cite{dimenet}). Similarly, the update of vector features is defined as:
\begin{align}
    \Delta \vec{h}_i = &\sum_{j \in \mathcal{N}(i)} \vec{h}_i \odot \phi_{vv}(s_j) \odot W_{vv}(||r_{ij}||) \\+  &\sum_{j \in \mathcal{N}(i)} \phi_{vs}(s_j) W_{vs}(||r_{ij}||) \frac{\vec{r}_j-\vec{r}_i}{||r_{ij}||},
\end{align}
where the first half of the equation is a convolution with respect to the vector features, and the second half is a convolution with respect to the scalar features using an equivariant filter. After the message passing block, the scalar and vector features are updated with the calculated residuals:
$s_i \leftarrow s_i + \Delta s_i, \quad \vec{h}_i \leftarrow  \vec{h}_i + \Delta \vec{h}_i$ and then fed into update block. Next, in the update block, the update of scalar features is calculated as:
\begin{equation}
    \Delta s_i = a_{ss} (s_i, ||W_v\vec{h}_i||) + a_{sv} (s_i, ||W_v\vec{h}_i||) <W_u\vec{h}_i, W_v\vec{h}_i>,
\end{equation}
where $a_{ss}$ and $a_{sv}$ are learnable function, and $W_v, W_u$ are learnable linear projection matrices. And then the update of vector features is calculated as:
\begin{equation}
    \Delta \vec{h}_i = a_{vv} (s_i, ||W_v\vec{h}_i||)W_u\vec{h}_i.
\end{equation}
In practice, $a_{ss}, a_{sv}, a_{vv}$ are derived from the same network $a$ by splitting along the feature dimension of output. The scalar features and vector features are again updated with residuals. Equivariant Transformer \cite{tholke2022torchmd} extends the above message passing layer to attention. GVP-GNN \cite{jing2021learning} leverages a similar idea with different design choices of the message passing layer. As demonstrated in Soledad Villar et al. \cite{villar2021scalars}, such formulation is expressive enough for approximating SE(3) equivariant functions. 

It is worth noting that the above models are just non-exhaustive instances of equivariant graph neural networks. There are several other lines of works that leverage different principles such as Lie algebra \cite{lieconv, lietransformer} , or other message passing protocols designed for vector features \cite{gemnet, spherenet, radialfield, huang2022equivariant} and more general groups \cite{egnn}.

\section{Graph Neural Networks on Molecular Applications}

GNNs have been widely implemented to various applications in molecular sciences \cite{mater2019deep, wieder2020compact, xiong2021graph}. This section reviews the molecular applications empowered by GNNs, including molecular property prediction, molecular scoring and docking, molecular dynamics simulation, molecular optimization and generation, and others. Each subsection introduces the formulation of the problem, prevalent datasets and metrics, as well as works to solve the problem using GNNs.

\begin{figure}[t]
    \centering
    \includegraphics[width=\linewidth]{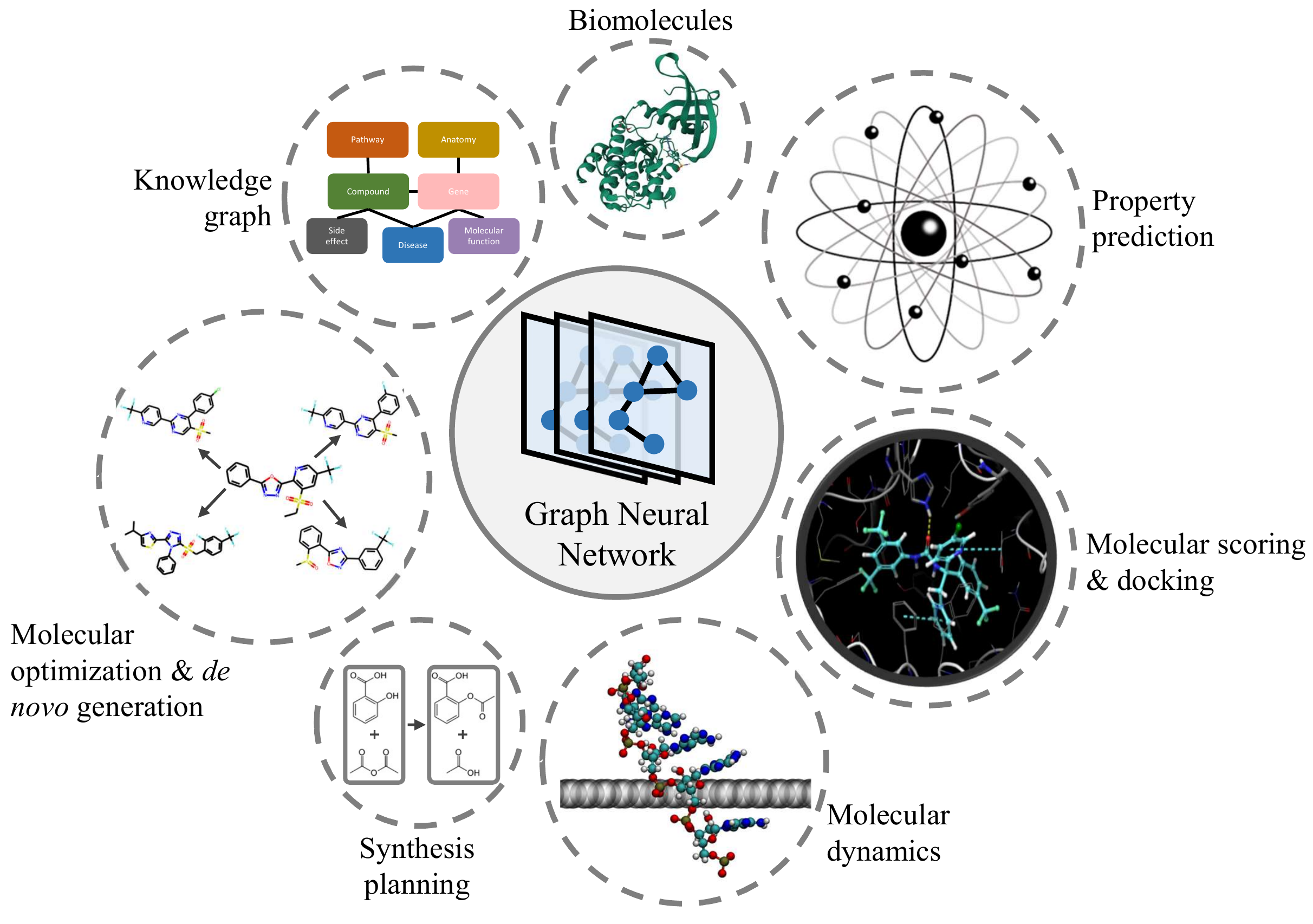}
    \caption{Various applications of GNNs on molecular sciences.\protect\footnotemark}
    \label{fig:applications}
\end{figure}

\subsection{Molecular Property Prediction}

The most straightforward utilization of GNNs is to predict the molecular properties given the molecule graphs \cite{wieder2020compact}. Such a task can be considered as a graph classification or regression following the empirical risk minimization in supervised learning. 
For a dataset of $N$ molecule graph and property pairs $\{(\mathcal{G}_1, y_1), \dots, (\mathcal{G}_N, y_N) \}$, the objective to optimize is given in Equation~\ref{eq:empirical_risk}. 
\begin{equation}
    \min_\theta \sum_i \ell (\text{GNN}_\theta(\mathcal{G}_i), y_i),
\label{eq:empirical_risk}
\end{equation}
where $\text{GNN}_\theta(\cdot)$ is a GNN model parameterized by $\theta$ to predict a certain property from an input molecule graph $\mathcal{G}_i$ and $\ell(\cdot, \cdot)$ measures the difference between the prediction and ground truth label, like cross-entropy loss for classification tasks and mean squared error for regression tasks \cite{coley2017convolutional, deng2021xgraphboost, aldeghi2022roughness}. Researchers have collected multiple databases containing a wide variety of molecular properties so that different GNN models can be benchmarked. 
\footnotetext{Resources of some fragments in Fig.~\ref{fig:applications}: \cite{cao2019water, schwaller2020predicting, wu2018moleculenet}, \url{https://www.dgl.ai/news/2020/06/09/covid.html}, \url{https://www.gla.ac.uk/news/archiveofnews/2021/september/headline_812517_en.html}, \url{https://www.deepmind.com/blog/alphafold-a-solution-to-a-50-year-old-grand-challenge-in-biology}.}
Table~\ref{tb:benchmarks} summarizes the domain, the number of compounds, the number of tasks, task type, whether containing 3D information and the sources of the popular databases for molecular property predictions. MoleculeNet \cite{wu2018moleculenet} is a widely used benchmark for molecular property prediction built upon multiple public databases. It contains multi-level properties, including physiology, biophysics, physical chemistry, and quantum mechanics. Besides, benchmarking-GNN \cite{dwivedi2020benchmarking} creates ZINC and AQSOL datasets for predicting the constrained and aqueous solubility. It should be pointed out that benchmarking-GNN only uses a subset of the original ZINC database (250,000 molecules) \cite{irwin2005zinc}. Recent efforts have focused on generating labeled molecular datasets via density functional theory (DFT) calculations. QM9, which is also included in MoleculeNet, contains approximately 134k molecules with their geometric, energetic, electronic, and thermodynamic properties. Alchemy \cite{chen2019alchemy} extends the previous dataset to include 12 quantum mechanical properties of 119,487 molecules with up to 14 heavy atoms. PCQM4M and its latest version PCQM4Mv2 in the OGB challenge aim at predicting the DFT-calculated HOMO-LUMO energy gap from 2D or 3D molecule graphs. ANI-1 \cite{smith2017ani} extends this concept of such datasets by including the energies of off-equilibrium conformations. Specifically, ANI-1 contains 24,687,809 conformations from 57,462 molecules. There are also works that create curated datasets for specific tasks, including quantitative structure-odor relationship (QSOR) \cite{sanchez2019machine}, Carbon absorption \cite{lei2014gas}, etc. Molecular property prediction is challenging since different datasets in it have numbers of instances of different magnitudes, from less than 200 to about 4,000,000 molecules. Also, most classification tasks in it have quite imbalanced labels, therefore, ROC-AUC and PRC-AUC are widely used instead of accuracy to measure the performance of different models. Regression benchmarks adapt either root-mean-square error (RMSE) or mean absolute error (MAE) as evaluation metrics. Besides, Hu et al. \cite{wu2018moleculenet} introduce scaffold splitting to separate structurally different molecules into training, validation, and test subsets. This strategy provides a more challenging yet realistic setting than random splitting and has been leveraged in many works on molecular property prediction \cite{yang2019analyzing, xiong2019pushing, Hu2020Strategies, rong2020self, wang2022molclr, fang2022geometry}. 

\begin{table}[t!]
  \centering
\resizebox{\textwidth}{!}{
  \begin{tabular}{llllllll}
    \toprule
    Domain & Dataset & \# Molecules & \# Tasks & Task type & 3D & Source \\
    \midrule
    \multirow{7}{*}{Physiology} & MUTAG & 188 & 1 & Classification & No & \cite{debnath1991structure} \\
    & SIDER & 1,427 & 27 & Classification & No & \cite{kuhn2016sider, wu2018moleculenet} \\
    & ClinTox & 1,478 & 2 & Classification & No & \cite{gayvert2016data, wu2018moleculenet} \\
    & BBBP & 2,039 & 1 & Classification & No & \cite{martins2012bayesian, wu2018moleculenet} \\
    & NCI1 & 4,110 & 1 & Classification & No & \cite{wale2008comparison} \\
    & Tox21 & 7,831 & 12 & Classification & No & \cite{2014tox21, wu2018moleculenet} \\
    & ToxCast & 8,575 & 617 & Classification & No & \cite{richard2016toxcast, wu2018moleculenet} \\
    \midrule
    \multirow{5}{*}{Biophysics} & BACE & 1,513 & 1 & Classification & No & \cite{subramanian2016computational, wu2018moleculenet} \\
    & CCRF-CEM & 3,047 & 1 & Regression & No & \cite{cortes2019kekulescope} \\
    & PC-3 & 4,294 & 1 & Regression & No & \cite{cortes2019kekulescope} \\
    & PDBBind & 19,443 & 1 & Regression & Yes & \cite{wang2004pdbbind, wu2018moleculenet, su2018comparative} \\
    & HIV & 41,127 & 1 & Classification & No & \cite{hiv, wu2018moleculenet} \\
    & MUV & 93,087 & 17 & Classification & No & \cite{rohrer2009maximum, wu2018moleculenet} \\
    & PCBA & 437,929 & 128 & Classification & No & \cite{wang2012pubchem, wu2018moleculenet} \\
    \midrule
    \multirow{5}{*}{\shortstack{Physical\\Chemistry}} & FreeSolv & 642 & 1 & Regression & No & \cite{mobley2014freesolv, wu2018moleculenet} \\
    & ESOL & 1,128 & 1 & Regression & No & \cite{delaney2004esol, wu2018moleculenet} \\
    & Lipophilicity & 4,200 & 1 & Regression & No & \cite{mendez2019chembl, wu2018moleculenet} \\
    & AQSOL & 9,982 & 1 & Regression & No & \cite{sorkun2019aqsoldb, dwivedi2020benchmarking} \\
    & ZINC & 12,000 & 1 & Regression & No & \cite{irwin2005zinc, dwivedi2020benchmarking} \\
    \midrule
    \multirow{7}{*}{\shortstack{Quantum\\Mechanics}} & QM7 & 7,165 & 1 & Regression & Yes & \cite{rupp, wu2018moleculenet} \\
    & QM7b & 7,211 & 14 & Regression & Yes & \cite{1367-2630-15-9-095003, wu2018moleculenet} \\
    & QM8 & 21,786 & 12 & Regression & Yes & \cite{ruddigkeit2012enumeration, wu2018moleculenet} \\
    & QM9 & 133,885 & 12 & Regression & Yes & \cite{ramakrishnan2014quantum, wu2018moleculenet} \\
    & Alchemy & 119,487 & 12 & Regression & Yes & \cite{chen2019alchemy} \\
    & PCQM4M & 3,803,453 & 1 & Regression & No & \cite{hu2020open} \\
    & PCQM4Mv2 & 3,378,606 & 1 & Regression & Yes & \cite{hu2020open} \\
    & ANI-1 & 24,687,809 (conf.) & 1 & Regression & Yes & \cite{smith2017ani} \\
    \bottomrule
  \end{tabular}}
  \caption{Summary of benchmarks for molecular property predictions.}
  \label{tb:benchmarks}
\end{table}

Most of the GNN architectures introduced in Section~\ref{sec:gnn} and \ref{sec:molecular_gnn} can be leveraged on molecular property predictions. Apart from those models, a wide range of GNN variants have been developed for the application. Wieder et al. \cite{wieder2020compact} provide a comprehensive survey for the molecular property predictions via GNNs till the year 2020. Early works utilize spectral GNNs for molecular property prediction \cite{henaff2015deep, li2018adaptive, liao2019lanczosnet, ma2019graph}. Since the proposal of GCN \cite{kipf2016semi}, a simple yet generic algorithm, many GNNs have been built upon it for molecular property predictions \cite{xu2017deep, li2017learning, wang2019molecule, cho2019enhanced, feinberg2020improvement}. Some works have also explored using RNNs for aggregation in this area \cite{altae2017low, ryu2018deeply, shindo2019gated}. After the formalization of the message-passing framework by Gilmer et al. \cite{gilmer2017neural}, researchers have built GNNs systematically following the insight to predict a wide range of molecular properties \cite{yang2019analyzing, withnall2020building, tang2020self, ma2020multi}. 

Recently, there are emerging researches that leverage DNNs, especially SE(3)-invariant or SE(3)-equivariant GNNs with 3D geometric information, for quantum mechanics (QM) predictions \cite{smith2017ani1, butler2018machine, zhang2018deep}. Many works are evaluated on QM9. Early SE(3)-invariant GNNs, including SchNet \cite{schutt2017schnet}, PhysNet \cite{unke2019physnet}, HIP-NN \cite{lubbers2018hierarchical}, MGCN \cite{lu2019molecular}, DimeNet \cite{gasteiger2019directional}, DimeNet++ \cite{klicpera2020fast}, Cormorant \cite{anderson2019cormorant}, SphereNet \cite{liu2021spherical}, are built upon interatomic distances and angles. Some works have leveraged orbital information to better predict QM properties \cite{qiao2020orbnet, karamad2020orbital, qiao2022informing, boiko2022stereoelectronics}. GNNs built upon SE(3)-equivariant operations have also been widely applied to this application. Examples include SE(3)-Transformer \cite{fuchs2020se}, E(n)-GNN \cite{satorras2021n}, LieTransformer \cite{hutchinson2021lietransformer}, L0/L1Net \cite{miller2020relevance}, GemNet \cite{gasteiger2021gemnet}, PaiNN \cite{schutt2021equivariant}, and TorchMD-Net \cite{tholke2022torchmd}. Many SE(3)-equivariant GNNs have also been benchmarked MD17 \cite{chmiela2017machine} which contains energy and force field predictions from molecular dynamics simulations of eight molecules. We will discuss GNN applications on MD simulations in Section~\ref{sec:md} extensively. Recent works have investigated \textit{ab-initio} solution of the Schr{\"o}dinger equation to acquire fundamental QM properties equipped with GNNs. Hermann et al. \cite{hermann2020deep} propose PauliNet, a GNN-based wave function ansatz. PauliNet adapts SchNet as the GNN backbone and exploits multireference Hartree-Fock solution as the optimization start point. The whole framework is trained using variational quantum Monte Carlo (VMC). Gao et al. \cite{gao2021ab} introduce a framework combining a GNN and a neural wave function to solve the Schr{\"o}dinger equation for multiple geometries via the VMC method simultaneously. 


\subsection{Molecular Scoring and Docking} 

Molecular scoring mainly refers to the prediction of pharmacological activity of a molecule candidate, which is essential in \textit{in silico} drug discovery \cite{xiong2021graph}. Such a method is widely applied in the virtual screening of drug candidates. It can also be leveraged to narrow down the search space for \textit{de novo} molecular generation. There are two major categories of molecular scoring: ligand-based scoring and structure-based scoring. The ligand-based scoring requires no information about the receptor in pharmacological activities. Such a setting can be considered as a special case of molecular property predictions that model the quantitative structure-activity relationships (QSAR), like drug-likeness, toxicity, solubility, etc. On the contrary, structure-based scoring is based on the structure of the biological target (usually a protein) of a molecule. Such a strategy predicts the drug-target interactions given the drug candidates and the target receptors. Molecular docking predicts the preferred pose of a molecule when bound with a binding site to form a stable protein-ligand complex. It is one of the most popular techniques in structure-based drug discovery. Several works leverage 3D CNN to model the protein-ligand complex \cite{stepniewska2018development, mcnutt2021gnina, yang2021deep, bao2021deepbsp, yadav2022prediction}. However, 3D CNN requires voxelization of 3D space which adds burdens to the memory. Moreover, such a voxelization loses resolution of the exact pose in 3D and lacks explicit modeling of interatomic interactions. This chapter reviews the structure-based molecular scoring and molecular docking empowered by GNNs. 

PDBBind \cite{wang2004pdbbind, wang2005pdbbind, liu2017forging} collects biomolecular complexes with their experimental binding affinity data and is widely used in structure-based scoring with deep learning as a regression task. DUD-E \cite{mysinger2012directory} is another dataset that contains 102 targets across different protein families. For each target, both positive (active) and negative (decoy) ligands are provided, which are formalized as classification tasks. Also, BindingDB \cite{liu2007bindingdb, gilson2016bindingdb} is a public database of measured binding affinities of proteins and ligands. Drugbank \cite{wishart2006drugbank, wishart2008drugbank}, a public online database, that collects FDA-approved drugs with drug target and drug action information. ChEMBL \cite{mendez2019chembl} also provides plenty of protein-ligand information including chemical compound and activity assay data. Several works have also created curated databases \cite{gao2018interpretable, zheng2020predicting}. 

Structure-based scoring aims at predicting the binding affinity of drug-target interactions (DTIs) \cite{lim2021review}. The key to designing GNN models for structure-based molecular scoring is how to integrate the information of the protein and/or the binding pocket to the ligand \cite{karimi2019deepaffinity, li2021machine, li2021structure}. Feinberg et al. \cite{feinberg2018potentialnet} introduce PotentialNet, a pioneering GNN framework that models the protein-ligand complexes. PotentialNet takes two stages in message-passing: (1) intra-molecular message-passing based only on covalent bonds, and (2) intra- and inter-molecular message-passing based on Euclidean distance and bonds. Gomes et al. \cite{gomes2017atomic} introduce a framework contain three parameter-sharing Siamese GNNs that encode ligand, protein, and protein-ligand complex, to features $G_\text{ligand}$, $G_\text{protein}$, and $G_\text{complex}$, respectively. The prediction of binding affinity then utilizes $\Delta G = G_\text{complex} - G_\text{ligand} - G_\text{protein}$. Lim et al. \cite{lim2019predicting} improve the framework by conducting intra-molecular message passing and inter-molecular massage-passing simultaneously. The model utilizes the subtraction between the features of a target protein and the ligand in their complex to predict the binding affinity. Such a strategy that models the interactions within the ligands and interactions between ligands and targets is widely used in later works. InteractionGraphNet \cite{jiang2021interactiongraphnet} by Jiang et al. models the intra- and inter-molecular interactions sequentially with two independent GNN models. Morrone et al. \cite{morrone2020combining} adapt a dual-GNN framework to encode the ligand graph and protein-ligand interactions separately. The output of the two GNNs is concatenated and fed to an MLP to predict the binding affinity of the complex. Son et al. \cite{son2021development} propose GraphBAR, a framework containing multiple graphs whose adjacency matrices cover neighbors within different distance cutoffs.  Knutson et al. \cite{knutson2022decoding} propose two parallel GNNs with one incorporating domain knowledge of proteins and ligands while the other learning interactions with no domain prior. The works listed above are built upon protein-ligand complexes. However, these models may suffer to make accurate predictions for those complexes without experimental measurements. To this end, Torng and Altman \cite{torng2019graph} propose to encode protein pocket graphs and 2D molecule graphs independently and combine the output features to predict the DTIs. Recent works have introduced language models to encode protein information in predicting DTIs \cite{gao2018interpretable, nguyen2021graphdta, moesser2022protein, wang2020gnn, wang2022structure, yang2022mgraphdta}. 

Unlike structure-based scoring which predicts binding affinity directly, molecular docking focuses on predicting the posture of a protein-ligand complex. Conventional docking contains a search module to generate massive potential protein-ligand binding poses and a scoring module to evaluate the interaction of binding poses \cite{lengauer1996computational, kitchen2004docking, ciemny2018protein}. Such methods rely on heavy candidate sampling and usually involve empirical scoring functions, thus may be time-consuming and inaccurate \cite{erickson2004lessons, wang2016comprehensive}. Recently, there are emerging works that apply GNNs for molecular docking \cite{ganea2021independent, garcia2021dockstring, jin2022antibody, shen2022boosting}. Jiang et al. \cite{jiang2022predicting} develop a GNN to refine the initial docking pose from conventional docking software. The GNN can be run multiple times to gradually get optimal predictions. Mendez et al. \cite{mendez2021geometric} report DeepDock, which first learns a statistical potential based on distance likelihood and then samples ligand conformations based on learned potential. Specifically, DeepDock models statistical potentials on torsion angles to generate favorable molecular conformations in the binding site \cite{klebe1994fast}. However, these methods assume prior knowledge of the binding site for the ligands. Some works have investigated predicting protein binding sites using GNNs \cite{fout2017protein, li2017drug, dai2021protein, sverrisson2022physics, zong2022beta}. To this end, GNN methods have been developed to both locate the binding site and predict the preferred binding conformations. EquiBind \cite{stark2022equibind} by St{\"a}rk et al. adapts SE(3) GNN to predict the binding site and applies constraints that only allow change of torsion angles while keeping bond lengths and angles fixed. TANKBind \cite{lu2022tankbind} by Lu et al. incorporates trigonometry constraints to GNNs which avoid unrealistic conformations like atom overlapping. It also leverages contrastive learning to build the energy landscape for the inter-molecular interaction of different binding sites. 


\subsection{Molecular Dynamics Simulation}
\label{sec:md}
Molecular Dynamics (MD) has a wide range of applications in material science, chemistry, and biophysics \cite{MD-for-all-2018, Karplus2002-MD-biomolecules, md-drug}. It provides a numerical way to study and predict intricate molecular systems. In essence, MD simulation calculates atomic forces and then updates the system states with a discretized equation of motion. The forces can either be modeled by \textit{ab initio} approaches (AIMD) like density functional theory (DFT) \cite{DFT-review-2014} that considers the electronic structure of atoms or through empirical potentials which bypasses electronic structures \cite{review-FFs-2018}. AIMD is highly accurate but its computation is prohibitively expensive, which limits its scope of application. On the contrary, empirical force fields are much more efficient, yet with worse accuracy and limited generalizability. The major limitation of MD simulations using empirical force fields stems from the difficulty in how to describe complex interatomic potentials accurately with the appropriate functional form given the diverse types of interactions in the system. In this regard, neural networks have the potential to close the accuracy gap between empirical potentials and AIMD with its function approximation capability \cite{Bartok-ML-modeling, ML-interatomic-2019,  review-Barati-2020, review-Noe-2020, GN-universal-molecules-2019}. 

Depending on how the environment of the atom is described, neural networks can be categorized into two broad categories. The first kind of approach relies on hand-designed featurization, which builds tailored descriptors that exploit domain knowledge \cite{Bartok-ML-modeling, ML-forcefield-Ab-initio-2017, smith2017ani1, FCHL-revisited-2020, DPMD-2018, fingerprints-o2}. One of the first examples is Behler-Parrinello Neural Networks (BPNN) \cite{Behler-2011-ACSF, Behler-NN-Energy-2007, Behler-BPNN2017}, which proposes atom-centered symmetry functions (ACSFs) for describing the local neighborhood around an atom. Graph neural networks, on the other hand, provide a framework to directly learn the atomic representations from the raw atomic coordinates and low-level atomic features, offering an alternative to the manually tailored atomic descriptors \cite{duvenaud2015convolutional, schutt2017quantum, unke2019physnet, schutt2017schnet, gilmer2017neural, dimenet, paiNN2021}. 

Based on the above frameworks, neural networks can be used to address the efficiency-accuracy tradeoff in MD from the following aspects. The first aspect is to use neural networks to learn the forcefield (i.e. interatomic potential) of a molecular system. In this case, neural networks trained on data with ab initio level accuracy can serve as fast and efficient surrogate models for AIMD. The output of the network can either be the potential energy surface (PES) or atomic forces, in turn forming an energy-based model (also known as neural network potentials) \cite{schutt2017schnet, chmiela2017machine, dimenet, Behler-BPNN2017, smith2017ani1, ml-potential-review, ml-energy-prl, schutt2017quantum, deep-potential, zubatyuk2019accurate} or force-based model \cite{forcenet, Mailoa2019, gnnff2021, gamd2022}. These two variants are similar in principle (Chmiela et al. \cite{chmiela2017machine} demonstrate that force-based model essentially learns a linearization of the PES), but force-based models are more accurate at predicting forces in practice and can bypass the process of calculating energy gradients. However, force-based models are generally not energy conserving and thus can generate unphysical predictions. Another perspective is learning the pattern of molecular trajectories instead of the dynamics, which treats the simulation as a sampling problem. Here neural networks are used as a probabilistic generative model \cite{score-md, coarse-score-md, dl-sampling-md, adaptive-dl-sampling} to model the distribution $p(x_{t+1}|x_t)$ with $x_{t}$ denotes the state of the system at time $t$. Since this paradigm no longer depends on dynamics to update the system, it does not suffer from the truncation error that arises from the discretization error of the equation of motion and thus can adopt a very large time step size (e.g. at the scale of a nanosecond). The downside is such a method can lead to unphysical prediction and cannot be used to study properties related to energy and dynamics.


\subsection{Molecular Optimization and Generation} 

Efficient and effective molecular generation is of crucial importance in practice \cite{elton2019deep}. Especially, in the pharmaceutical industry, the discovery of new drugs is a long and expensive process that costs more than \$2.5 billion and 10–15 years on average \cite{dimasi2016innovation}. Therefore. it is appealing to develop techniques that automatically and effectively generate plausible molecule candidates. The deep generative models, which learn to approximate the distribution of observed data, have been leveraged for molecule graph optimization and \textit{de novo} generation \cite{yang2019concepts, dimitrov2019autonomous, schneider2020rethinking, jimenez2020drug, xiong2021graph, xie2022advances}. The optimization of the molecule usually starts with a hit compound and manipulates the graph to achieve better target properties like toxicity,  drug-likeness, binding affinity, etc. The \textit{de novo} molecule generative models generate novel molecules from scratch or conditioned on desired properties or specific fragments. In this case, molecule optimization can be considered as a conditional generation problem conditioned on the hit molecule. Thus, we review these two applications together in this section. There are two major frameworks in molecule graph generation: (1) fragment-based generative models and (2) molecule-based generative models \cite{sun2020graph, xiong2021graph}. The fragment-based models generate molecule graph iteratively, where at each step the model choose an action, including adding, deleting, or editing one or multiple atoms, bonds, or functional groups. Such a fragment-based framework can be incorporated with reinforcement learning (DRL) that defines a Markov decision process (MDP) and learns the best generative policy by maximizing the expected accumulated rewards. It can also be implemented with autoregressive models like RNNs, which determine the action based on statuses at previous steps \cite{popova2019molecularrnn, bongini2021molecular}. The molecule-based model, on the other hand, generates all the attributes (i.e., atoms and bonds) of a molecule at once. Many generative methods equipped with GNNs fall into this category, including variational autoencoder (VAE) \cite{kingma2013auto}, generative adversarial network (GAN) \cite{goodfellow2014generative}, flow-based generative model \cite{rezende2015variational}, and score-based or diffusion generative model \cite{song2019generative, ho2020denoising}. VAE usually contains an encoder and a decoder that are trained to maximize the evidence lower bound concerning the log-likelihood of training data. GAN introduces two jointly trained components: a generator and a discriminator, where the generator generates samples while the discriminator attempts to distinguish between samples from the training data and those from the generator. Flow-based models are built upon a sequence of invertible operations that directly models the likelihood of the observed data. Score-based or diffusion model defines a Markov chain that adds random noise to data at each step and learns to denoise from the noise to recover the data. The following part in this section introduces molecule graph generative methods in detail. 

\subsubsection{Metrics and Benchmarks}

To measure the performance of deep generative models on molecule graphs, one needs to define comprehensive metrics and benchmarks \cite{brown2019guacamol, polykovskiy2020molecular}. Simple but effective metrics include validity which assesses whether the generated molecules are valid, uniqueness/diversity which evaluates if a model generates a different molecule at each sampling, and novelty which measures whether generated molecules exist in the training set. Besides, Preuer et al. propose the Fr{\'e}chet ChemNet distance (FCD) \cite{preuer2018frechet} following the widely-used Fr{\'e}chet Inception distance (FID) \cite{heusel2017gans} in image synthesis, which is built upon the difference of hidden representations from ChemNet, a deep neural network, between generated molecules and those in the training set. Kullback−Leibler (KL) divergence is also used to measure whether the generative model approximates the distribution of the training set. In conditional generation or optimization, certain properties are used to evaluate the generative models. Many works build evaluation metrics on drug-likeness \cite{gomez2018automatic}. For example, the synthetic accessibility score (SAS) describes the ease of synthesis of molecules based on fragment contributions and a complexity penalty \cite{ertl2009estimation}, the octanol-water partition coefficient $\log P$ characterizes the drug-likeness of a molecule, the penalized $\log P$ is the subtraction of original $\log P$ and SAS, and quantitative estimate of drug-likeness (QED) applies desirability functions which provides a multicriteria metric to assess drug-likeness \cite{bickerton2012quantifying}. Maximum mean discrepancy (MMD) \cite{gretton2012kernel, you2018graphrnn} is another metric to evaluate the generated graphs. MMD is employed to measure the distribution difference between generated molecule graphs and training set on degrees, clustering coefficients, orbit counts, as well as bond lengths of different types. For candidate drug optimization or generative, the binding affinity of molecules with respect to the target protein pocket computed by molecular docking tools \cite{friesner2004glide, trott2010autodock} or molecular dynamic (MD) simulations \cite{eastman2017openmm, hollingsworth2018molecular, cournia2020rigorous} are also used as the objective \cite{jeon2020autonomous, eckmann2022limo}. Energy-related properties, like HOMO-LUMO gap and dipole moment, have also been leveraged as the optimization or generation targets \cite{hoogeboom2022equivariant}. Additionally, Gao et al. \cite{gao2022sample} point out that sampling efficiency should be another important consideration in real molecular generation applications. 

Several works have contributed to creating molecular generation benchmarks and databases. GuacaMol \cite{brown2019guacamol} by Brown et al. employ a standardized subset from ChEMBL database \cite{mendez2019chembl}. It includes validity, uniqueness, novelty, FCD, and KL divergence as the evaluation measurements. MOSES \cite{polykovskiy2020molecular} by Polykovskiy et al. contains 1,936,962 molecular structures selected from ZINC \cite{irwin2005zinc} and is split into the training, test, and scaffold test sets containing approximately 1.6M, 176k, and 176k structure, respectively. Aside from the metrics included on GuacaMol, MOSES implements the similarity of fragments and scaffolds and properties distribution to evaluate the generative models. Besides, ZINC \cite{irwin2005zinc}, ChEMBL \cite{mendez2019chembl}, GDB databases \cite{blum2009970, ruddigkeit2012enumeration} are also used in training 2D molecule graph generative models. QM9 \cite{ramakrishnan2014quantum} which contains more than 130K 3D molecular structures, is also widely used for 3D molecular generation. Recently, GEOM \cite{axelrod2022geom} by Axelrod et al. introduces GEOM-QM9, an extension to QM9, containing multiple conformations for most molecules, and GEOM-Drugs containing 304,466 drug-like species up to a maximum of 91 heavy atoms. 

\subsubsection{Fragment-based Generative Model}

Fragment-based generative model modifies a molecule graph by adding, removing, or substituting a fragment (i.e., an atom or a motif) sequentially \cite{grover2019graphite}. Such a framework can be directly modeled as an RL problem \cite{olivecrona2017molecular, guimaraes2017objective, zhavoronkov2019deep, zhou2019optimization, jeon2020autonomous, wang2021efficient}. At each time step $t$, an agent receives a reward $r_t$ and predicts an action $a_t \in \mathcal{A}$ given the current state $s_t \in \mathcal{S}$, and the next state $s_{t+1} \in \mathcal{S}$ is depended solely on $s_t$ and $a_t$ following the MDP setting. Here, $\mathcal{A}$ and $\mathcal{S}$ denote the action and state space, respectively. In this case, the state is an incomplete molecule graph, the action is what fragment to add, remove, or substitute, and the agent is usually modeled by a deep neural network, including GNNs. Besides, the reward $r_t$ evaluates how well the molecule is generated which could be a chemical metric or an output from property prediction ML models that approximate empirical measurements. The optimization objective of an RL agent (Equation~\ref{eq:rl}) is to maximize the expected accumulated return $R_t$ of a policy defined by the agent. 
\begin{equation}
    \mathbbm{E}_\pi [R_t] = \mathbbm{E}_\pi \left[ \sum_{t}^T \gamma^t r_t \right], 
\label{eq:rl}
\end{equation}
where $\gamma \in (0,1]$ is a discount rate and $T$ is the maximal length of a trajectory. RL provides a generic framework for molecular design of different targets \cite{grebner2020automated}. You and Liu et al. \cite{you2018graph} propose a graph convolutional policy network (GCPN) that successively constructs a molecule by adding an atom, a substructure, or a bond, and is trained via the policy gradient algorithm. It built its agent on GCN \cite{kipf2016semi}. Experiments show that GCPN can generate molecules with optimized or targeted penalized $\log P$ or QED properties. Jin et al. \cite{jin2020multi} present RationaleRL composed of rationale extraction, graph completion, and rationale distribution. RationaleRL is evaluated on GNK3$\beta$ and JNK3 that measures the inhibition against glycogen synthase kinase-3 c-Jun N-terminal kinase-3 \cite{li2018multi}. It also measures the validity, diversity, and novelty of the generative model. DeepGraphMolGen proposed by Khemchandani et al. \cite{khemchandani2020deepgraphmolgen} adapts GNNs for action prediction as well as property prediction in an RL setting. It designs a multi-objective reward function containing SAS, QED, and predicted binding affinity at the dopamine and norepinephrine transporters. Besides RL, autoregressive models can be incorporated with fragment-based molecule graph generation \cite{popova2019molecularrnn, mercado2021graph}. GraphINVENT by Mercado et al. \cite{mercado2021graph} leverages a tiered GNN architecture to generate a single bond at each step. Podda et al. \cite{podda2020deep} and Chen et al. \cite{chen2021deep} develop autoregressive generative models that operate on the fragments. Lim et al. \cite{lim2020scaffold} propose to extend a given molecule scaffold by sequentially adding nodes and edges, which guarantees the generated molecules preserve the certain scaffold. Xie et al. \cite{xie2021mars} propose a framework named MARS which combines molecule graph editing with MCMC sampling. The method employs a GNN to sample fragment-editing actions at each step adaptively. Shi et al. \cite{shi2020graphaf} introduce GraphAF built upon an autoregressive normalized flow-based generative model that adds nodes and edges to a molecule graph sequentially. GraphAF can also be fine-tuned with RL for molecular optimization of certain properties. Following the normalized flow model, GraphDF by Luo et al. \cite{luo2021graphdf} samples discrete latent variables which are mapped to additional nodes and edges via invertible modulo shift transforms.

Fragment-based generative models introduced so far have demonstrated the effectiveness of generating valid, unique, and diverse molecules as well as generating or optimizing towards desired SAS, $\log P$, and QED. However, these models only adapt 2D topological information without 3D geometric features. Many recent works have investigated generative models on 3D molecular structures as many properties, like energy and protein-ligand binding, are related to 3D geometries. Gebauer et al. \cite{gebauer2019symmetry} introduce G-SchNet, an autoregressive generative model for 3D molecular structure generation via placing atoms in 3D Euclidean space one by one. Gebauer et al. further introduce cG-SchNet \cite{gebauer2022inverse}, a conditional version of G-SchNet to generate molecules with certain motifs or low-energy (e.g., small HOMO-LUMO gap). Simm et al. \cite{simm2020reinforcement} propose an RL formulation that adds atoms in 3D euclidean space sequentially and designs a reward function based on the fast quantum-chemical calculated electronic energy. The team also introduce MolGym, an RL environment comprising several molecular design tasks along with baseline models. Flam-Shepherd et al. \cite{flam2022scalable} further introduce a 3D molecule generative RL framework to add fragments instead of single atoms at each step, which is more efficient in creating larger and complex molecules. Luo et al. \cite{luo2021autoregressive} present G-SphereNet follows the flow-based generation while determining atoms in 3D space by predicting distances, angles, and torsion to preserve equivariance. Besides, researchers have investigated generating 3D molecules given a designated protein binding site. Docking scores along with other generic metrics are utilized to evaluate the deep generative models.  Luo et al. \cite{luo20213d} and GraphBP by Liu et al. \cite{liu2022generating} develop 3D autoregressive generative methods to sample the type and position of a new atom sequentially in the 3D binding pocket. Powers et al. \cite{powers2022fragment} propose a molecular optimization method that expands a small fragment into a larger molecule within a protein site. At each step, a GNN trained by imitation learning selects the connecting point as well as the type and dihedral angle of a fragment to be connected. To preserve the equivariance in the 3D molecular generation, many works adopt a local spherical coordinate system in the generation process \cite{gebauer2019symmetry, gebauer2022inverse, simm2020reinforcement, luo20213d, liu2022generating}. Besides, Imrie et al. \cite{imrie2020deep} introduce DeLinker which generates linkers given two fragments in 3D Euclidean space.

\subsubsection{Molecule-based Generative model}

Unlike fragment-based generative models, molecule-based generative models create the node features, edge features, and adjacency matrix of a molecule graph simultaneously. VAE, a popular generative model, has been leveraged for molecule-based generation \cite{blaschke2018application, gomez2018automatic}. A VAE is composed of two sub-models: an encoder that encodes the input into a latent feature domain, and a decoder that maps the feature back to the input \cite{kingma2013auto, kipf2016variational}. The loss for training a VAE is given in Equation~\ref{eq:vae}. 
\begin{equation}
    \ell = - \mathbbm{E}_{z \sim q_\theta(z|x)}[\log p_\phi (x'|z)] + \text{KL}(q_\theta(z|x) \| p(z)),
\label{eq:vae}
\end{equation} 
where $x$ is a input data, $z$ is the latent vector, $q_\theta$ is the encoder, and $p_\phi$ is the decoder. The first term measures how well the model reconstructs the data and the second term regularizes the latent vector to be similar to a prior Gaussian distribution $p(z)$ through a KL divergence. Such a framework has been leveraged to generate molecule graphs \cite{kipf2016variational, kusner2017grammar, vignac2022topn}. Simonovsky et al. \cite{simonovsky2018graphvae} propose GraphVAE, where the encoder is modeled by a GNN and the decoder is modeled by an MLP that outputs the probability of node features, edge features, and adjacency matrix given the predefined number of nodes. The reconstruction loss measures the difference between generated and input molecule graphs in terms of node features, edge features, and adjacency matrix. Kwon et al. \cite{kwon2019efficient} improve the framework with an approximate graph matching for efficient reconstruction loss. Besides, the work incorporates RL and an auxiliary property prediction to improve molecule graph generation. Ma et al. \cite{ma2018constrained} investigate regularized VAE to generate semantically valid molecule graphs. Constraints included in the decoder include atomic valence, graph connectivity, and node label compatibility. Bresson et al. \cite{bresson2019two} design a graph VAE with GCN \cite{kipf2016semi} as the encoder and a novel two-step decoder. In the decoder, an MLP first predicts the molecular formula from the latent vector and then a GCN learns to place bonds between atoms from the same latent vector. Jin et al. \cite{jin2018junction} propose JT-VAE, a two-phase generative model employing subgraphs. JT-VAE first generates a junction tree (JT) representing the scaffold of chemical substructures and then the substructures are assembled into a complete molecule graph. Jin et al. \cite{jin2020hierarchical} further improve the framework by leveraging larger and more flexible motifs and a decoder that operates in a hierarchical coarse-to-fine manner. The model demonstrates better performance in larger molecule generation compared to previous works. Li et al. \cite{li2019deepscaffold} present DeepScaffold that generates molecules based on scaffolds. DeepScaffold contains DeepScaffold a VAE with a GNN encoder to complete atom and bond types for scaffold, a scaffold-based generator, and a filter for pharmacophore constraints. Mahmood et al. \cite{mahmood2021masked} also report masked molecular generation which learns the conditional distribution of unobserved atoms and bonds given observed components. VAE-based model can also be directly leveraged for conditional generation with respect to targeted properties \cite{kang2018conditional, lim2018molecular, lee2022mgcvae, richards2022conditional}. Besides, given the learned latent representation by the VAE framework, molecular optimization can also be conducted in the latent space \cite{griffiths2020constrained, samanta2020nevae, hoffman2022optimizing, chenthamarakshan2020cogmol, ragoza2022generating}. Jin et al. \cite{jin2018learning} report VJTNN, a molecular optimization framework based on JT-VAE, which includes stochastic latent codes to capture meaningful molecular variations. To avoid infeasible optimized molecules, the model introduces an adversarial training method to align the distribution of graphs on the latent domain. VJTNN achieves competitive performance on QED, penalized $\log P$, and biological activity optimization. LIMO, by Eckmann et al. \cite{eckmann2022limo}, adapts VAE with an inceptionism-like reverse optimization to optimize generated molecules towards better binding affinity. 

Apart from the VAE-based model, GAN has also been leveraged for molecule graph generation \cite{wang2019learning, guo2022deep}. GAN \cite{goodfellow2014generative, mirza2014conditional, arjovsky2017wasserstein, karras2019style} learns to approximate the distribution of actual data $p_x(x)$ through a min-max game between a generator $G$ and a discriminator $D$. The discriminator is trained to classify real samples from generated fake ones, while the generator learns to generate samples from random noise $z \sim p_z(z)$ that the discriminator fails to recognize. The objective of GAN is given in Equation~\ref{eq:gan}. 
\begin{equation}
    \min_G \max_D \left( \mathbbm{E}_{x \sim p_{x}(x)} [\log D(x)] + \mathbbm{E}_{z \sim p_{z}(z)} [\log (1 - D(G(z))] \right), 
\label{eq:gan}
\end{equation}
where $G$ attempts to generate plausible samples by minimizing the objective while $D$ attempts to better detect fake samples by maximizing the objective. Cao et al. \cite{de2018molgan} implement the architecture of GAN to molecule graph generation and propose MolGAN. The generator in MolGAN generates an adjacency matrix and feature matrix that combine to define a molecule graph. Despite the generator and discriminator as in the standard GAN, MolGAN adapts an RL agent to generate molecules with targeted chemical properties (e.g., QED, SAS, $\log P$). Maziarka et al. \cite{maziarka2020mol} introduce Mol-CycleGAN for molecular optimization, following CycleGAN \cite{zhu2017unpaired}. Mol-CycleGAN generates a structurally similar molecule with preferred properties given a start compound. The model demonstrates effectiveness in optimizing penalized $\log P$. Tsujimoto et al. \cite{tsujimoto2021molgan} further present L-MolGAN, which improves MolGAN performance on large molecule generation by penalizing the disconnected generated molecule graphs. 

Other generative models have also been investigated for the molecule graph generation. The normalized flow-based model can also be leveraged in molecule-based generation besides fragment-based generation \cite{liu2019graph}. 

A flow-based model defines a invertible deterministic transformation $f_\theta$ parameterized by $\theta$ between the data space $X$ and the latent space $Z$ as $f_\theta: Z \rightarrow X$ \cite{rezende2015variational, kobyzev2020normalizing}. $Z$ follows a prior Gaussian distribution $p_Z$, the log-likelihood of a data $x$ is given in Equation~\ref{eq:flow}.
\begin{equation}
    \log p_X(x) = \log p_Z (f_\theta^{-1}(x)) + \log | \det J |,
\label{eq:flow}
\end{equation}
where $J = \frac{\partial f_\theta^{-1}(x)}{x}$ is the Jacobian matrix of the function $f_\theta^{-1}(x)$. The objective of training a flow-based model is to maximize the log-likelihood. To efficiently compute $\det J$, the affine coupling mapping is leveraged in a normalized flow-based model \cite{dinh2014nice, dinh2016density}. GraphNVP by Madhawa et al. \cite{madhawa2019graphnvp} and MoFlow by Zang et al. \cite{zang2020moflow} employ reversible normalizing flow-based model for the molecule graph generation. Both models decompose the generation into two phases, namely the generation of the adjacency matrix and node attributes, which yields the complete molecule graphs. Score-based \cite{song2019generative, song2020improved} or diffusion generative models \cite{ho2020denoising, song2020denoising} are also investigated in molecule graph generation \cite{niu2020permutation}. Song et al. \cite{song2020score} propose a unified framework that generalizes score-based or diffusion generative models through stochastic differential equations (SDEs). The forward diffusion process that transforms data $x$ to a simple noise distribution is given in Equation~\ref{eq:sde}. 
\begin{equation}
    dx = f(x, t)dt + g(t) dw, 
\label{eq:sde}
\end{equation}
where  $w$ is the standard Wiener process, $f(\cdot, t)$ is the drift coefficient of $x(t)$, and $g(\cdot)$ is the diffusion coefficient of $x(t)$. To sample data from random noise, the reversed SDE is given in Equation~\ref{eq:reversed_sde}. 
\begin{equation}
    dx = [f(x,t) - g^2(t) \nabla_x \log p_t(x)] dt + g(t) d \tilde{w}, 
\label{eq:reversed_sde}
\end{equation}
where $\tilde{w}$ is a standard Wiener process during the reversed flow. The training objective for a score function $s_\theta(x, t)$ parameterized by $\theta$ such that $s_\theta(x, t) \approx \nabla_x \log p_t(x) $ is given in Equation~\ref{eq:score}.
\begin{equation}
    \mathbbm{E}_{t \sim \mathcal{U}(0,T)} \mathbbm{E}_{p_t(x)} [ \lambda(t) \| \nabla_x \log p_t(x) - s_\theta (x) \|_2^2 ], 
\label{eq:score}
\end{equation}
where $\lambda(t)$ is a positive weighting function and $\mathcal{U}(0, T)$ is a uniform distribution over the diffusion time interval. Hoogeboom et al. \cite{hoogeboom2022equivariant} propose equivariant diffusion model (EDM) which is built upon DDPM \cite{ho2020denoising}, a diffusion generative framework, to generate 3D molecules with an equivariant GNN model. EDM treats molecules as atomic types and coordinates without explicitly considering interatomic bonds. The framework has been demonstrated effective for random and conditional generation. Besides, diffusion model has been introduced to biomolecule generation (e.g., proteins) with GNNs \cite{trippe2022diffusion}.

\subsubsection{Molecular Conformation Generation}

Despite molecular optimization and generation via GNN and deep generative models, another important application of generative models is molecular conformation generation, which aims at predicting the ensemble of low-energy 3D conformations of a molecule from its 2D graph solely \cite{axelrod2020molecular, xu2021molecule3d}. Understanding the low-energy conformations, the most stable configurations of molecules in 3D Euclidean space, is of great importance. Since 3D molecular structures determine the functions of chemical and biological processes \cite{alquraishi2021differentiable}. Popular metrics for molecular conformation generation include the matching score (MAT) and the coverage score (COV) \cite{hawkins2017conformation, shi2021learning, xu2022geodiff} which are both built up root-mean-square deviation (RMSD) computing the normalized Frobenius norm of the discrepancy of two aligned atomic coordinate systems \cite{kabsch1976solution}. Formally, let $S_g$ denote the sets of generated conformations and $S_r$ denote the one with reference conformations. The expression of MAT and COV are given in Equation~\ref{eq:mat} and \ref{eq:cov}, respectively.
\begin{equation}
    \text{MAT}(S_g, S_r) = \frac{1}{|S_r|} \sum_{\mathbf{R} \in S_r} \min_{\mathbf{\hat{R}} \in S_g} \text{RMSD}(\mathbf{R}, \mathbf{\hat{R}}),
\label{eq:mat}
\end{equation}
\begin{equation}
    \text{COV}(S_g, S_r) = \frac{1}{|S_r|} |\{ \mathbf{R} \in S_r | \text{RMSD}(\mathbf{R}, \mathbf{\hat{R}}) < \delta, \mathbf{\hat{R}} \in S_g \}|,
\label{eq:cov}
\end{equation}
where $\delta$ is a pre-defined threshold. Generally, a lower
MAT score indicates a better accuracy and a higher COV score indicates a better diversity for the generative model. In many cases, $\delta$ is set as $0.5 \angstrom$ for QM9 and $1.25 \angstrom$ for GEOM-Drugs. Ganea et al. \cite{ganea2021geomol} further propose recall metrics, MAT-R and COV-R, to measure the number of correctly predicted conformers, as well as the precision metrics, MAT-P and COV-P, to measure the number of generated structures of high quality. 

Mansimov et al. \cite{mansimov2019molecular} proposed to leverage a VAE framework with MPNN to directly predict the atomic coordinates from 2D graphs, which does not conserve the SE(3)-equivariance. To this end, the following works instead predict invariant geometric attributes like interatomic distances and torsional angles. Simm et al. \cite{simm2019generative} propose graph distance geometry (GraphDG) following the VAE framework to predict the interatomic distances. Conditional graph continuous flow (CGCF) by Xu et al. \cite{xu2021learning} predicts the distance matrix via a flow-based model and optimizes generated conformations via a Markov chain Monte Carlo (MCMC) process and an energy-based tilting model (ETM). Both GraphDG and CGCF utilize the distance geometry (DG) method \cite{liberti2014euclidean} to search atomic coordinates from the predicted distance matrix. ConfVAE by Xu et al. \cite{xu2021end} improves the VAE pipeline which learns to encode molecule graphs into latent space and computes the 3D conformations as a principled bilevel optimization problem. However, these methods predict the distances and conformations separately where the errors in predicted distances accumulate in conformation calculations. In some cases, the predicted distance matrix may even fail to preserve valid 3D conformations. To avoid the issues caused by the two-stage strategy, ConfGF by Shi et al. \cite{shi2021learning} and DGSM by Luo et al. \cite{luo2021predicting} directly estimates the gradient fields of the logarithm density of atomic coordinates (pseudo force fields) via a score-based generative method. The former relies on static molecule graphs to predict conformations as previous methods, while the latter deploys a dynamic graph construction that better models long-range interactions. Further, GeoDiff by Xu et al. \cite{xu2022geodiff} leverages the denoising diffusion probabilistic model (DDPM), a diffusion generative model, to generate 3D conformations via reversing a diffusion process. To keep the SE(3)-equivariance, ConfGF, DGSM, and GeoDiff leverage equivariant GNNs to make predictions of atomic coordinates. Besides interatomic distances and atomic positions, recent works have investigated to predict torsional angles in conformation generation. Ganea et al. propose GeoMol \cite{ganea2021geomol} which predicts local 3D structures and ensembles the local structures by predicting the torsion angles. Jing et al. \cite{jing2022torsional} introduce torsional diffusion, a diffusion generative model that formalizes molecular conformation generation in the space of torsional angles. TorsionNet by Gogineni et al. \cite{gogineni2020torsionnet} presents the conformation generation as a reinforcement learning problem to sample torsion at each time step. 

In summary, various generative models, including VAE, GAN, flow-based model, and score-based/diffusion model, have been implemented for a molecular generation. The fragment-based generation defines the generative process as sequentially adding atoms, bonds, or motifs to build the graph. Such a framework, incorporated with RL or autoregressive models, makes the process tractable. RL provides a suitable framework for molecular optimization. Recently works that utilize a flow-based model to generate molecules autoregressively have also been a success. The molecule-based generation, on the other hand, generates the whole molecule simultaneously. Within this category, VAE-based generative models are widely used and explicitly learn a meaningful latent space where molecular search and optimization can be easily conducted. A few works leverage GAN models for molecular generation, however, it is not as widely used as VAE, due to potentially unstable training and hard to manipulate the generation. Flow-based models can also be leveraged for the molecule-based generation which generates node features, bond features, and/or adjacency matrices simultaneously. Recently, the development of score-based or diffusion models in image generation has also inspired works in molecular generation. Moreover, recent research is paying more attention to 3D molecular structure generations beyond 2D graphs. Since 3D structures play an essential role in various molecular applications. Notably, there are also massively works that build molecular generative models on string-based representations (e.g., SMILES) \cite{kadurin2017drugan, gomez2018automatic, segler2018generating, prykhodko2019novo}. However, this is not the focus of this chapter, and compared to string-based representations, molecule graph generation with GNNs provides more topological and geometric information that is vital in molecular science. 


\subsection{Others}

\subsubsection{Synthesis Planning and Retrosynthesis Prediction} 

Synthesis planning aims at determining the synthesis path of a chemical compound from available starting materials through a series of chemical reactions \cite{coley2017prediction}. Retrosynthetic prediction, formalized by E. J. Corey \cite{corey1967general}, refers to the reverse problem: to select suitable disconnections recursively given the desired product \cite{coley2018machine}. The major challenge in the computational retrosynthetic analysis is the combinatorial exploration space of chemical reactions for target compounds. There are two major approaches: template-based and template-free methods. The former matches the target molecule to chemical reaction rules to yield one or multiple candidate precursors. The latter directly predicts the reactants from target products. Recently, several works have investigated template-free or semi template-free retrosynthesis prediction via GNNs. USPTO-50k \cite{schneider2016s} which contains 50k atom-mapped reactions with 10 reaction classes is the most widely used benchmark. G2G proposed by Shi et al. \cite{shi2020graph} learns to translate a target molecule into a set of reactant molecule graphs. G2G contains two steps: (1) edit prediction that splits the target molecule into multiple synthons via predicting the reaction centers, and (2) synthon completion that transforms the synthons into reactant graphs. Such a workflow has been leveraged by many works. Sun et al. \cite{sun2020energy} propose to formalize such a retrosynthesis prediction problem as an energy-based model (EBM). Somnath et al. \cite{somnath2021learning} introduce leaving graphs from a precomputed vocabulary for synthon completion, which greatly reduces the complexity of synthon generation. Lin et al. \cite{lin2022g2gt} further present G2GT, which leverages recent graph transformer (i.e., Graphormer \cite{ying2021transformers}) in template-free retrosynthesis prediction. 
Han et al. \cite{han2022gnn} propose GNN-Retro that combines GNN with the A* search algorithm. Besides retrosynthetic prediction, GNNs have been implemented other tasks concerning chemical reactions, including predicting reaction products \cite{do2019graph, sacha2021molecule}, reaction conditions \cite{ryou2020graph, ryou2020graph}, reaction yields \cite{saebi2021graph, kwon2022uncertainty}, and synthesizability \cite{bradshaw2019model, gu2022perovskite, liu2022retrognn}. 

\subsubsection{Molecular Knowledge Graph}

Knowledge graphs (KGs) are graphical structured knowledge bases where each node models a knowledge entity and each edge models the relations between entities \cite{miller1995wordnet, dumontier2014bio2rdf, lehmann2015dbpedia}. KGs are a powerful tool that incorporates multiple data sources. In molecular science, KGs have been leveraged to represent biomedical knowledge. Specifically, nodes may model various entities, including drugs, protein targets, diseases, side effects, pathways, etc. Such biomedical KGs can be leveraged to predict drug-drug interactions (DDIs), which aims at predicting the outcomes of combined use of two or more drugs, like side effects, adverse reactions, or even toxicity \cite{ryu2018deep, niu2019pharmacodynamic, karim2019drug, celebi2019evaluation, zhang2022graph}. Sources of DDI knowledge include DrugBank \cite{wishart2006drugbank, wishart2018drugbank}, KEGG \cite{kanehisa2000kegg, kanehisa2021kegg}, BioSNAP \cite{biosnapnets}, PharmGKB \cite{whirl2012pharmacogenomics, whirl2021evidence}, Bio2RDF \cite{belleau2008bio2rdf, dumontier2014bio2rdf}, etc. In DDI KG, nodes model drugs and edges model interactions, thus DDI prediction can be treated as link prediction tasks. GNNs have since made progress on DDI prediction based on KGs. Feng et al. \cite{feng2020dpddi} implement GCN \cite{kipf2016semi} on drug-drug interaction KG to predict potential DDIs. Yu et al. \cite{yu2021sumgnn} introduce SumGNN which utilizes a local subgraph in KG around drug pairs instead of the entire KG to obtain information. Multiple works investigate leveraging multi-modal data to enhance DDI predictions \cite{bai2020bi, chen2021muffin}. Lyu et al. \cite{lyu2021mdnn} propose MDNN containing KG and heterogeneous pathways to obtain multi-modal representations. Lin et al. \cite{lin2020kgnn} propose an end-to-end framework built upon GNNs which captures potential neighborhoods of drugs by leveraging their relations in KG. Zhang et al. \cite{zhang2022mkge} propose MKGE which combines KG embedding with molecular structure features for DDI predictions. He et al. \cite{he2022multi} propose to fuse the topological information from molecular graphs and the interaction information in the form of SMILES. Some works directly build GNNs on molecular graphs of drug pairs to predict DDIs  \cite{feng2022prediction, nyamabo2021ssi, nyamabo2022drug} instead of relying on KGs. Besides DDI KGs, other KGs have also been investigated like protein-protein interaction KGs \cite{keskin2008prism, zitnik2019evolution, Hu2020Strategies, yang2020graph} and chemical reaction networks \cite{garay2022chemical}. Zitnik et al. \cite{zitnik2018modeling} propose Decagon based on graph autoencoder to predict side effects of drug pairs and build a large multi-modal KG including protein-protein, drug-protein, and drug-drug interactions. 

\subsubsection{Biomolecules}

Though the focus of this chapter lies in small organic molecules. It is worth mentioning that GNNs have also been leveraged to learn expressive representations of large biomolecules (i.e., proteins, RNAs, DNAs, etc.). AlphaFold \cite{jumper2021highly}, which is a milestone in protein structure prediction, utilizes attention and message-passing mechanisms. Following this, multiple works have probed improving protein structure prediction from amino acid sequences \cite{baek2021accurate, mirdita2022colabfold, lin2022language, wu2022high}. Many works have also developed GNN-based models in various protein applications, including protein function prediction \cite{spalevic2020hierarchical, gligorijevic2021structure, zhang2022protein}, protein-compound interactions \cite{evans2021protein, wang2021protein, yan2020graph, bryant2022improved} and protein design \cite{jing2020learning, ingraham2019generative, strokach2020fast, trippe2022diffusion, anand2022protein}. Besides proteins, several works have investigated RNAs and DNAs via GNNs \cite{townshend2020atom3d, wang2021scgnn, townshend2021geometric}.


\section{Self-supervised Learning on Molecule Graphs}

The previous section reviews various GNN architectures with expressive message-passing operations for molecular systems. However, a well-designed neural network does not guarantee to perform well, as the performance of the model is also closely related to the quantity and quality of data. In supervised learning, the ML models require instance-label pairs to learn representations. Labeling data can be expensive and time-consuming, and acquiring labels in molecular science takes even more effort. Measuring the attributes and properties of molecules depends on complicated wet-lab experiments and/or expensive simulations. Thus, labeled molecular data is far from sufficient in many applications. Besides, molecular labels are usually noisy since different experimental or simulation environments can lead to significant fluctuations. Moreover, the chemical space of potential molecules is magnificent. The number of potential pharmacologically active molecules is estimated to be in the order of $10^{23}$ to $10^{60}$ \cite{reymond2012enumeration}. For the GNN trained on limited and noisy data in a supervised learning manner, it is of great challenge to generalize novel molecules from the huge chemical space. Despite the limited labeled molecular data, the number of molecules in publicly available databases is growing fast in recent years. Such databases include ZINC \cite{irwin2005zinc}, PubChem \cite{wang2012pubchem}, Enamine REAL library, etc. A question may be asked: can one make use of the large unlabeled data to improve ML models? 

Following the insight, self-supervised learning (SSL) \cite{hadsell2006dimensionality,doersch2017multi} is proposed to utilize the large unlabelled data and learn better representations that are generalizable to various challenging applications. As the name indicates, SSL trains the ML models via the supervisory signals from the data itself, which usually utilizes the underlying structure in the data. There are two major categories of SSL: (1) non-contrastive learning methods, (2) contrastive learning methods. Many non-contrastive learning methods rely on generative models that learn meaningful representations by recovering the data from partially observed input. Examples in computer vision include colorization of grayscaled images \cite{zhang2016colorful}, prediction of rotation of images \cite{gidaris2018unsupervised}, and reconstruction of partially masked image \cite{pathak2016context, zhang2017split, bao2022beit, he2022masked}. In natural language processing (NLP), self-supervised pre-training via prediction of masked tokens has been a standard technique in deploying large language models \cite{devlin2018bert, liu2019roberta}. On the other hand, contrastive learning (CL) methods, like MoCo \cite{he2020momentum}, SimCLR \cite{chen2020simple}, and SwAV \cite{caron2020swav}, learns representations by attracting the representations of the positive pairs together and pushing representations of the negative pairs away, where positive pairs are perturbed instances from the same data while negative samples are instances from different data \cite{bengio2021deep}. For a query representation $\pmb q$ and a batch of $N+1$ keys containing only one key $\pmb k_+$ from a positive instance, the information noise-contrastive estimation (InfoNCE) loss, a popular type of contrastive loss, is given in Equation~\ref{eq:infonce}.
\begin{equation}
    \ell = - \log \frac{\exp((\pmb q^\top \pmb k_+) / \tau)}{\sum_{i=0}^{N} \exp((\pmb q^\top \pmb k_i) / \tau)},
\label{eq:infonce}
\end{equation}
where $\tau$ is the temperature parameter. Such a loss can be considered as a $(N+1)$-way classifier to classify $\pmb q$ as its positive instance $\pmb k_+$. CL has been a success in a wide variety applications \cite{xie2020pointcontrast, chaitanya2020contrastivemedical, gao2021simcse, you2020graphcl, magar2022crystal}. Following the objective of contrasting representation in CL, researchers design non-contrastive methods that only require pulling together representations of positive pairs, like BYOL \cite{grill2020bootstrap}, SimSiam \cite{chen2021exploring}, Barlow Twins \cite{zbontar2021barlow}, and VICReg \cite{bardes2021vicreg}, which simply the framework. 

\begin{figure}[t]
    \centering
    \includegraphics[width=0.8\linewidth]{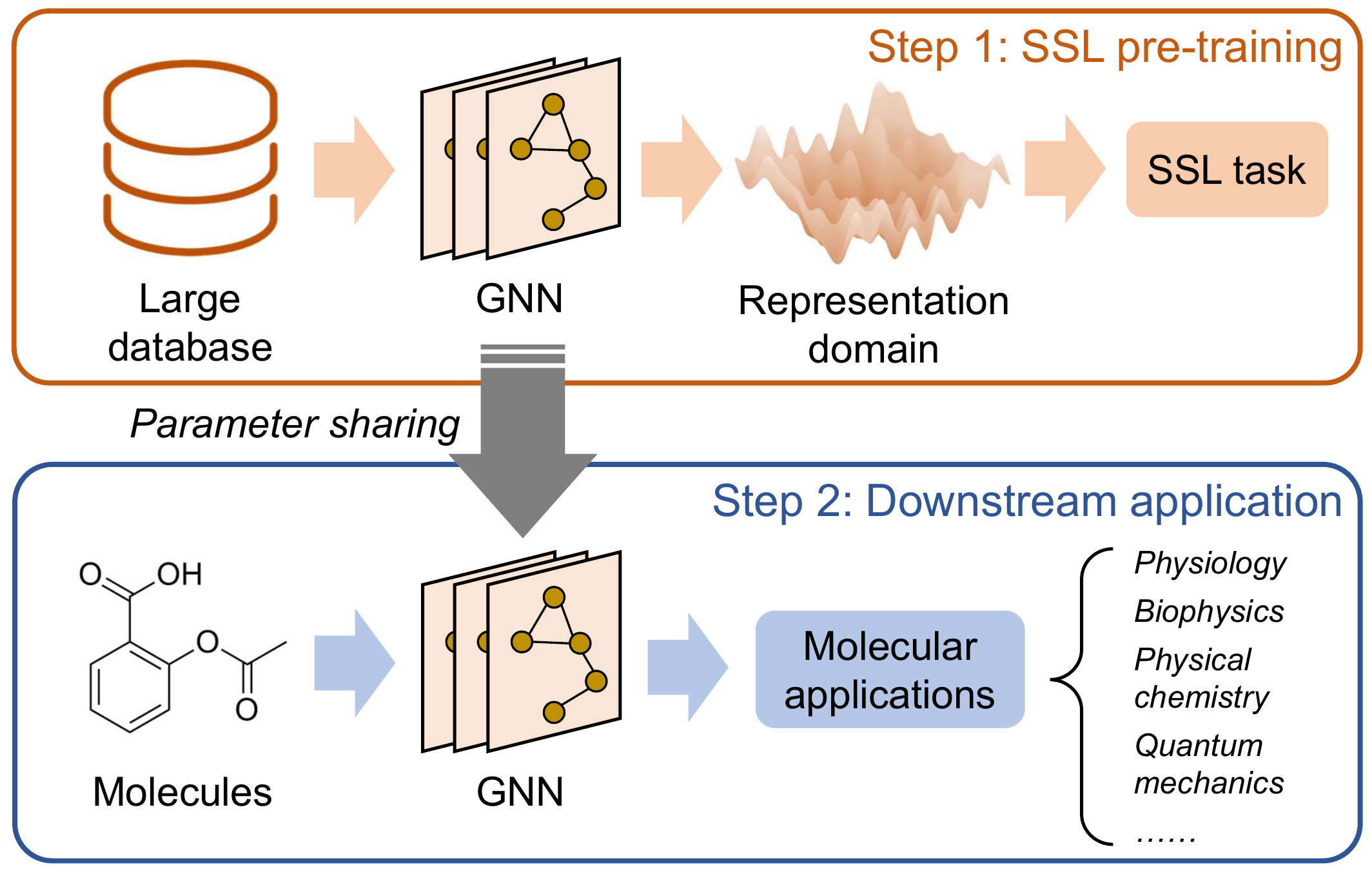}
    \caption{Illustration of self-supervised learning pre-training and fine-tuning on downstream tasks.}
    \label{fig:ssl}
\end{figure}

Since proposed, SSL has been introduced to learning molecular representations from large unlabeled data with GNNs \cite{xie2022self}. As shown in Figure~\ref{fig:ssl}, SSL on molecule graphs has two steps. In the pre-training step, the GNN is trained on large unlabeled molecular data via self-supervised learning tasks, where the GNN parameters are optimized to encode expressive molecular representations. In the downstream application step, the pre-trained GNN is fine-tuned or used to directly extract the learned representations for various molecular applications. By this means, the GNN model digs into the molecule graphs and learns better representations that can perform well even on tasks with noisy and limited data. The following paragraphs will enumerate works that investigate different SSL techniques on molecule graphs.

Non-contrastive SSL methods are first introduced to learn representations from 2D molecule graphs. Most of such molecular SSL methods are based on predicting the probability or attribute of a masked or partially observed substructure occurring within a molecule \cite{fang2022knowledge}. Hu et al. \cite{Hu2020Strategies} introduce pre-training tasks at the levels of both individual nodes and the whole graphs. The node-level pre-training contains two self-supervised tasks: context prediction which uses subgraphs to predict their surrounding graph structures, and masked attribute prediction which pre-trains GNNs through predicting the attributes of masked nodes/edges. While the graph-level pre-training is based on supervised learning of domain-specific graph labels, which is still constrained by the availability of molecular labels. Rong et al. \cite{rong2020self} further introduce contextual property prediction which predicts masked subgraphs and graph-level motif prediction which predicts the containing functional groups within a molecule as two SSL tasks. This work also adapts a transformer-based model to learn representations from molecule graphs. Zhang et al. \cite{zhang2021motif} also design motif-level self-supervised tasks that leverage the BRICS algorithm to construct motif trees and pre-trains GNNs by generating molecular graphs motif-by-motif. At each step of pre-training, the model first predicts the topology which determines whether a node has extra motifs to be connected to, and then predicts the motif label from motif vocabulary. He et al. \cite{he2022mmm} extends the idea of predicting masked attributes even further by leveraging information in chemical reactions. The work named masked molecule modeling (MMM) adapts Graphormer \cite{ying2021transformers} to take in molecule graphs in a chemical reaction, the GNN model is then trained to reconstruct randomly masked atoms, motifs, and bonds within the reaction. Besides, Liu et al. \cite{Liu2019NGramGS} leverage N-gram graph representations to embed molecule graphs in an unsupervised manner that assembles the node features in short walks. The graph embedding method requires no training and can be used for different ML models. Sun et al. \cite{sun2019infograph} introduce InfoGraphn which learns presentations by maximizing the mutual information between the graph-level embedding and multi-level substructure embedding. 

Apart from 2D graphs, recent works have probed SSL strategies to include 3D geometric information. To acquire 3D conformations for pre-training, some works \cite{fang2022geometry, li2022geomgcl, zhou2022uni} uses efficient Merck molecular force field (MMFF94) function in RDKit \footnote{\url{https://www.rdkit.org/}} to simulation 3D atomic positions. While other methods \cite{liu2021pre, stark20213d, zaidi2022pre, liu2022molecular, jiao20223d} leverage datasets containing 3D molecular conformations at equilibrium calculated by more precise yet expensive density functional theory (DFT), including QM9, GEOM, PCQM4Mv2, and Molecule3D. Chen et al. \cite{chen2021algebraic} design an algebraic graph-assisted bidirectional transformer method that fuses representations from the algebraic graph and transformer. The algebraic graph embeds 3D stereochemical information into graph invariants and the bidirectional transformer embeds representations from SMILES \cite{weininger1988smiles} language. The fused representations can be incorporated with a variety of machine learning algorithms, including tree-based models and DNNs, for various molecular property predictions. Fang et al. \cite{fang2022geometry} propose a geometry-enhanced molecular representation learning method (GEM) that learns representations through predicting bond lengths, bond angles, and interatomic distances. It also introduces an auxiliary graph-level SSL task that predicts molecular fingerprints, i.e., ECFP and MACCS keys. Zhou et al. \cite{zhou2022uni} present masked atom prediction and 3D position denoising including pair-distance prediction and direct coordinate prediction in the framework named Uni-Mol. The input molecule is perturbed by masking some atoms and adding noise to 3D coordinates randomly. Uni-Mol also adapts a transformer-based model with invariant spatial positional encoding to embed geometric information. With the recent development of equivariant GNNs on 3D molecules, researchers have explored SSL strategies that perform to improve the expressiveness of such models. Zaidi et al. \cite{zaidi2022pre}, Liu et al. \cite{liu2022molecular}, and Jiao et al. \cite{jiao20223d} have demonstrated the effectiveness of position denoising as self-supervised pre-training on equivariant GNNs. The three works share the similar idea that adds noise to the 3D atomic positions at equilibrium and let the GNN predict the noise \cite{zaidi2022pre, jiao20223d}, or predict the original interatomic distances \cite{liu2022molecular}. They have also shown that the positional denoising SSL objective is equivalent to the force field prediction from equilibrium structures. This sheds a light on leveraging equivariant GNNs for a wider range of applications like predicting potential energy surfaces and molecular dynamics simulations. 

Inspired by the success of CL, CL-based self-supervised methods have been introduced to molecular systems. Wang et al. \cite{wang2022molclr} propose MolCLR, a CL framework for molecular representation learning with GNNs. To create contrastive pairs, MolCLR implements three molecule graph augmentation strategies: atom masking, bond deletion, and subgraph removal. A positive pair contains graphs augmented from the molecule while a negative pair contains those augmented from different molecules. MolCLR adapts the normalized temperature-scaled cross entropy (NT-Xent) loss from SimCLR \cite{chen2020simple}, which is a modification of the InfoNCE loss. Given a batch of $N$ molecules, each molecule generates two augmented graphs, and the NT-Xent loss for a positive pair $(i,j)$ in the batch of $2N$ graphs is given in Equation~\ref{eq:ntxent}. 
\begin{equation}
    \ell_{i,j} = - \log \frac{\exp((\pmb z_i^\top \pmb z_j) / \tau)}{\sum_{k=1}^{2N} \mathbbm{1}_{\{k \neq i\}} \exp((\pmb z_i^\top \pmb z_k) / \tau)},
\label{eq:ntxent}
\end{equation}
where $\mathbbm{1}_{\{\cdot\}}$ is an indicator function, and $\pmb z_i$, $\pmb z_j$ are normalized representations of the positive pair. Zhang et al. \cite{zhang2020motif} further propose MICRO-Graph which extracts informative subgraphs from molecules via GNNs and performs contrastive pre-training on sampled subgraphs learned by GNNs. Similar to non-contrastive learning methods, CL has also gone beyond 2D molecule graphs and applies multi-view CL pre-training. Liu et al. \cite{liu2021pre}, St{\"a}rk et al. \cite{stark20213d}, and Li et al. \cite{li2022geomgcl} propose GraphMVP, 3D infomax, and GeomGCL, respectively. All three methods conduct contrastive pre-training between 2D topological graphs and 3D geometric structures to learn molecular representations with 3D information embedded. Namely, 2D and 3D views from a molecule compose a positive pair while others are negative pairs. GraphMVP also utilizes a generative model in the representation domain as an auxiliary SSL task. Besides, Zhu et al. \cite{zhu2021dual} develop a dual-view CL framework that contrasts between SMILES strings and 2D molecule graphs, where the former is encoded by a transformer and the latter is encoded by GNN. Further, MEMO from Zhu et al. \cite{zhu2022featurizations} presents a multi-view CL that leverages four molecular representations, including 2D topology, 3D geometry, SMILES string, and fingerprint. Some works have sought to fuse chemical domain knowledge in molecular CL models to learn better representations. Following the insight, Wang et al. \cite{wang2022improving} introduce iMolCLR to improve previous CL frameworks in two aspects: (1) mitigating faulty negative instances via considering molecular similarities in NT-Xent loss, and (2) fragment-level contrasting between substructures decomposed via the BRICS algorithm. Fang et al. \cite{fang2021knowledge} propose a molecule cluster strategy that utilizes functional groups and fingerprints similarity to select similar molecules as positive pairs and dissimilar molecules as negative pairs in contrastive training. The work also includes functional group embeddings as an input to GNN. MoCL by Sun et al. \cite{sun2021mocl} leverages local-level and global-level in contrastive learning on molecule graphs. Specifically, the local-level domain knowledge augments the molecule graph by replacing a substructure with a bioisostere which perturbs the instance while maintaining the chemical properties. The global-level domain knowledge introduces a loss that minimizes the difference between the similarity of learned representations and ECFP similarity between two molecules. Fang et al. \cite{fang2022molecular} propose knowledge-enhanced CL that includes graph augmentations based on the chemical element knowledge graph as well as the know-aware message-passing functions. In another model named CoSP, Gao et al., \cite{gao2022cosp} leverage information besides small organic molecules to learn better representations. CoSP adapts a cross-domain CL framework that generates contrastive pairs from ligand molecules and protein binding pockets. It also introduces ChemInfoNCE loss to reduce the negative sampling bias through a chemical similarity-enhanced negative ligand sampling strategy.


\section{Conclusion}

Recent years have witnessed an immense growth of graph neural networks in molecular sciences. In this survey, we review the generic message-passing framework for building graph neural networks. Graph neural network architectures designed for small organic molecules are introduced. Works on learning representations from 2D topological graphs as well as recent efforts in leveraging 3D geometric information are covered. Also, we try to provide a comprehensive overview of benchmarks, metrics, and graph neural network models in various molecular applications, including property predictions, molecular scoring, and docking, molecular optimization and generation, molecular dynamics, synthesis planning, molecular knowledge graph, etc. The last section epitomizes molecular self-supervised learning with graph neural networks. Graph neural networks have been successfully implemented to solve a wide variety of molecular tasks. However, the acquisition of high-quality labeled molecular data is still expensive and time-consuming. One direction that is worth exploring is to introduce domain knowledge to molecular graph neural networks for solving challenging problems with limited or noisy data. 



\printbibliography

\end{document}